\documentclass[9pt,journal,compsoc]{IEEEtran}
\ifCLASSOPTIONcompsoc
  \usepackage[nocompress]{cite}
  \usepackage{float}
\else
  \usepackage{cite}
\fi
\ifCLASSINFOpdf
\else
\fi
\newcommand\MYhyperrefoptions{bookmarks=true,bookmarksnumbered=true,
pdfpagemode={UseOutlines},plainpages=false,pdfpagelabels=true,
colorlinks=true,linkcolor={black},citecolor={black},urlcolor={black},
pdftitle={Bare Demo of IEEEtran.cls for Computer Society Journals},%<!CHANGE!
pdfsubject={Typesetting},%<!CHANGE!
pdfauthor={Michael D. Shell},%<!CHANGE!
pdfkeywords={Computer Society, IEEEtran, journal, LaTeX, paper,
             template}}%<^!CHANGE!
\usepackage[\MYhyperrefoptions,pdftex]{hyperref}
\usepackage{graphicx}
\usepackage{dsfont}
\usepackage{bbm}
\usepackage{bm}
\usepackage{cite}
\usepackage{picinpar}
\usepackage{amsmath}
\usepackage{url}

\usepackage{hyperref}
\hypersetup{hidelinks,
	colorlinks=true,
	allcolors=black,
	pdfstartview=Fit,
	breaklinks=true}
\usepackage{xcolor,colortbl}
\usepackage{flushend}
\usepackage[latin1]{inputenc}
\usepackage{colortbl}
\usepackage{caption}
\usepackage{subcaption}
\usepackage{soul}
\usepackage{multirow}
\usepackage{pifont}
\usepackage{color}
\usepackage{alltt}
\usepackage{enumerate}
\usepackage{siunitx}
\usepackage{epstopdf}
\usepackage{pbox}
\usepackage{amssymb}
\usepackage{mathrsfs}
\usepackage{tabularx}
\usepackage{booktabs}
\usepackage[ruled,vlined]{algorithm2e}
\begin{document}
% \title{Bare Advanced Demo of IEEEtran.cls for\\ IEEE Computer Society Journals} Weakly Supervised Segmentation SS3D: Semi-supervised Robust Framework  Data Region-Level Boundary Awareness and Instance Discrimination with Knowledge Distillation and Region-Level Boundary and Instance Discrimination with Label Generalized Label-Efficient 3D Scene Parsing via Hierarchical Feature Aligned Pre-Training and Region-Aware Fine-tuning Generalized Label-Efficient 3D Scene Parsing via Hierarchical Feature Aligned Pre-Training and Region-Aware Fine-tuning  and Region-Aware Fine-tuning Robot Hierarchical Generalized Scene Parsing  with Fast Rendering for
\title{Generalized Robot 3D Vision-Language Model with Fast Rendering and Pre-Training Vision-Language Alignment}
% Aligned Pre-LaminatedHierarchical training with Pre-trainingAlignment Alignment   Feature Aligned Pre-Training hierarchical al Data 
%  Hierarchical  Weakly-Supervised awareness\title{Generalized Label-Efficient 3D Scene Parsing with Knowledge Distillation and Region-Level Boundary Awareness and Instance Discrimination} Generalized Label-Efficient 3D Scene Parsing via Hierarchical Feature Aligned Pre-Training Generalized Label-Efficient 3D Scene Parsing via Hierarchical Feature Aligned Pre-Training

\author{Kangcheng Liu, \IEEEmembership{Member, IEEE}, Yong-Jin Liu, \IEEEmembership{Senior Member, IEEE}, and
Baoquan Chen, \IEEEmembership{Fellow, IEEE}

\IEEEcompsocitemizethanks
{

\IEEEcompsocthanksitem 
The preliminary results of this work are published at the European Conference on Computer Vision 2022 (ECCV 2022). 

\IEEEcompsocthanksitem Kangcheng Liu is with The Division of Engineering and Applied Science, California Institute of Technology (Caltech), Pasadena, USA.

\IEEEcompsocthanksitem 
Yong-Jin Liu is with the BNRist, MOE-Key Laboratory of Pervasive Computing, Department of Computer Science and Technology, Tsinghua University, Beijing, China. 
% National Key Lab of Genaral AI, Peking University, China; and  nd 

% \IEEEcompsocthanksitem 

\IEEEcompsocthanksitem 
Baoquan Chen is the National Key Lab of Genaral AI, Peking University, Beijing, China, and with the School of Artificial Intelligence, Peking University, Beijing, China.

% Department of Computing, Hong Kong Polytechnic
% University, Hung Hom, Kowloon, Hong Kong.\IEEEcompsocthanksitem 
% Chang Wen Chen is with the Department of Computing, Hong Kong Polytechnic
% University, Hung Hom, Kowloon, Hong Kong.

}% <-this % stops a space with the  IEEE HKUST April 30 August 26 6XX XX  E-mail: Changwen.chen@polyu.edu.hk. November 17,
% October Manuscript received January XX, 2023.
\thanks{Manuscript received January XX, 2023.}}
% November revised XX XX, 2023.  Kangcheng Liu and Ben M. Chen are with the Department of Mechanical and Automation Engineering, The Chinese University of Hong Kong, Shatin, HK 999077, China. (email: kcliu@mae.cuhk.edu.hk, benmec@cuhk.edu.hk) September
% space somehow managed to creep in. , ~Vol.~XX, No.~XX, XXXX~2022

% The paper headers Journal of \LaTeX\ Class Files
% \markboth{IEEE Transactions on Pattern Analysis and Machine Intelligence, ~Vol.~XX, No.~XX, XXXX~2022}%, 2025
\markboth{IEEE Transactions on Pattern Analysis and Machine Intelligence, 2025}%
{Shell \MakeLowercase{\textit{et al.}}: Bare Advanced Demo of IEEEtran.cls for IEEE Computer Society Journals}
\IEEEtitleabstractindextext{
\begin{abstract}
Deep neural network models have achieved remarkable progress in 3D scene understanding while trained in the closed-set setting and with full labels. However, the major bottleneck for the current 3D recognition approach is that these models do not have the capacity to recognize any unseen novel classes beyond the training categories in diverse real-world applications. In the meantime, current state-of-the-art 3D scene understanding approaches primarily require a large number of high-quality labels to train neural networks, which merely perform well in a fully supervised manner. Therefore, we are in urgent need of a framework that can simultaneously be applicable to both 3D point cloud segmentation and detection, particularly in the circumstances where the labels are rather scarce. This work presents a generalized and straightforward framework for dealing with 3D scene understanding when the labeled scenes are quite limited. To extract knowledge for novel categories from the pre-trained vision-language models, we propose a hierarchical feature-aligned pre-training and knowledge distillation strategy to extract and distill meaningful information from large-scale vision-language models, which helps benefit the open-vocabulary scene understanding tasks. To leverage the boundary information, we propose a novel energy-based loss with boundary awareness benefiting from the region-level boundary predictions. 
To encourage latent instance discrimination and to guarantee efficiency, we propose the unsupervised region-level semantic contrastive learning scheme for point clouds, using confident predictions of the neural network to discriminate the intermediate feature embeddings at multiple stages. In the limited reconstruction case, our proposed approach, termed \textit{WS3D++}, ranks 1st on the large-scale ScanNet benchmark on both the task of semantic segmentation and instance segmentation. Also, our proposed \textit{WS3D++} achieves state-of-the-art data-efficient learning performance on the other large-scale real-scene indoor and outdoor datasets S3DIS and SemanticKITTI. Extensive experiments with both indoor and outdoor scenes demonstrated the effectiveness of our approach in both data-efficient learning and open-world few-shot learning. The code is at: \href{https://drive.google.com/drive/folders/1M58V-PtR8DBEwD296zJkNg_m2qq-MTAP?usp=sharing}{\textit{WS3D++} Code link}.

\end{abstract}

% Note that keywords are not normally used for peer-reviewed papers. Weakly Supervised/Semi-Supervised Learning
}

% Make the title area
\maketitle

% not be used in make title, but will appear (i.e., to be "transported")
% is not selected <OR> if conference mode is selected - because compsoc
% conference papers position the abstract like regular (non-compsoc)
\IEEEdisplaynontitleabstractindextext
% compost under a non-conference mode.

% \fi
%
% For peer-review papers, this IEEEtran command inserts a page break and
% creates the second title. It will be ignored for other modes. approach
\IEEEpeerreviewmaketitle

\ifCLASSOPTIONcompsoc

\IEEEraisesectionheading{\section{Introduction}\label{sec:introduction}}
\else

\section{Introduction}
\label{sec:introduction}
\fi

% \vspace{-2.18mm}

\begin{figure}[ht]
\centering
\includegraphics[width=\linewidth]
{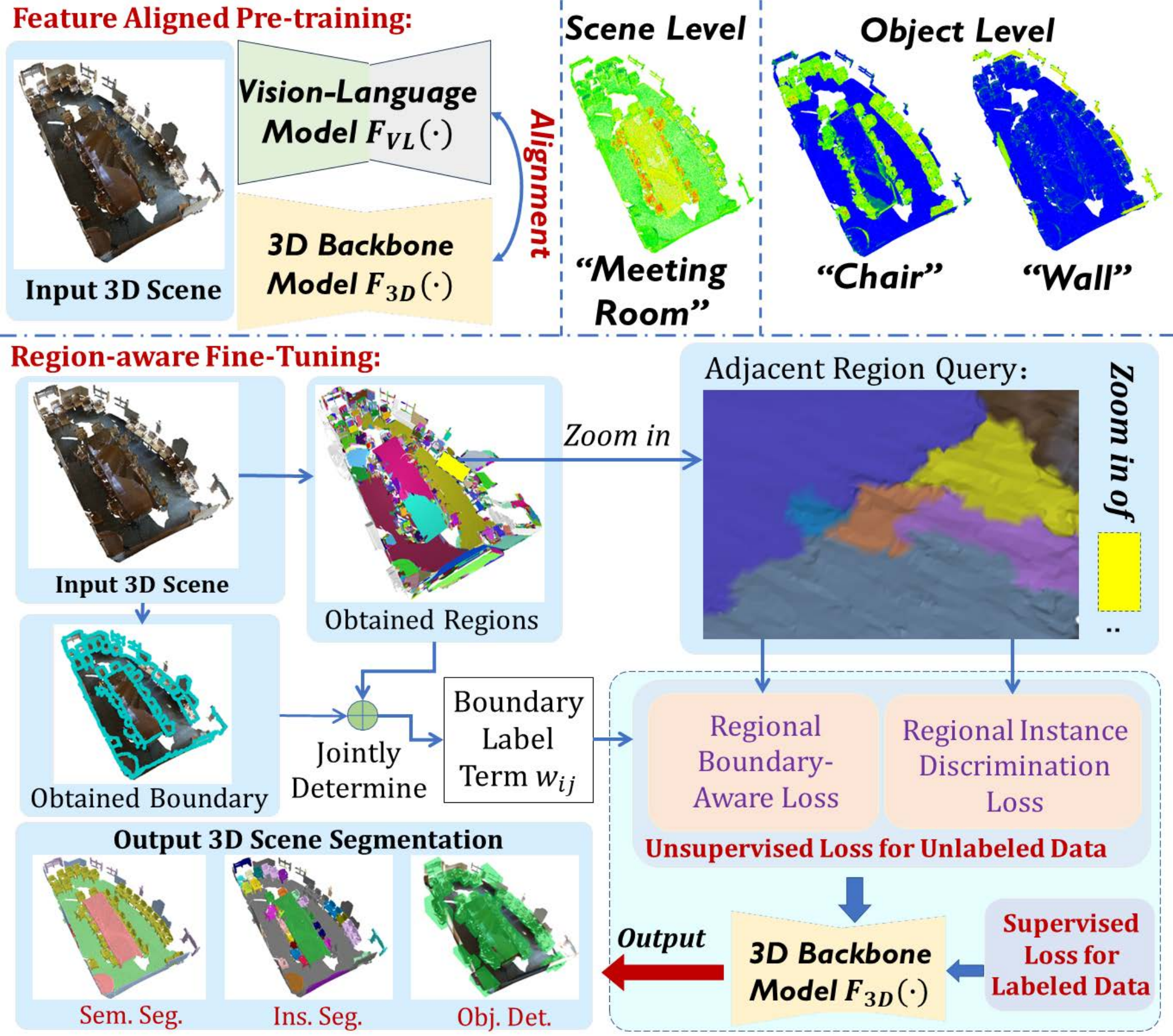}
\caption{The final overall illustrative diagram of our proposed \textit{WS3D++}. We integrate language-3D feature associated pre-training and data-efficient fine-tuning as a general scene parsing vision-language model to achieve effective \textit{data-efficient} as well as \textit{open-vocabulary} 3D scene understanding for 3D scenes.}
\label{Intro}
\vspace{-8mm}
\end{figure}
% \vspace{-8mm}\vspace{-8mm}\vspace{-12mm}\vspace{-7mm}\vspace{-7mm}\vspace{-2mm}\vspace{-7mm}\vspace{-7mm}\vspace{-6mm}\vspace{-1mm}\vspace{-5.3mm}\vspace{-5.3mm}-2.3mm\vspace{}5.3 \vspace{-1mm} \vspace{-1mm}\vspace{-1mm}
% \vspace{-3mm} \vspace{-5mm} \vspace{-2mm} \vspace{-2mm} \vspace{-2mm} \vspace{-2mm} \vspace{-2mm}

\IEEEPARstart{T}{he} 3D scene parsing problem, which typically encompasses several important downstream tasks: point cloud semantic segmentation, instance segmentation, and object detection, becomes increasingly important with the wide deployment of 3D sensors, such as LiDAR and RGB-D cameras~\cite{behley2021towards, ao2023buffer, zhang2023growsp, liu2022weakly, song2022ogc, rozenberszki2023unscene3d}. Point clouds are raw sensor data obtained from 3D sensors and the most simple and common 3D data representation for understanding 3D scenes of robot navigation, robot grasping, and manipulation tasks. 
% However the  robotics ly adopted  and Therefore, the success of current scalable and training and testing data   strides and  scene the  common 3D recently \textbf{}  in autonomous driving, augmented reality, and  of robot navigation, grasping,  and manipulation tasks
% \vspace{-0.0mm}
Despite significant success in deep neural networks applied to 3D visual perception, two major challenges hinder the construction of more scalable visual perception systems in 3D worlds. One is the \textbf{closed-set assumption}, which means the model only performs well while recognizing the categories that appear in the training set and struggles in recognizing the novel unseen object categories or concepts. Another is the heavy \textbf{reliance on large amounts of high-quality labeled data}. Large-scale 3D scenes are very laborious to label, which also makes it very hard for deep network models to perform well with very limited annotations.

% \includegraphics[width=\linewidth] \includegraphics[scale=0.26228] H

% \vspace{-3mm}
 % \vspace{-0.93006mm}  \vspace{-5mm} 5
 % \vspace{-3mm}  \vspace{-5mm}
% \vspace{-0.93006mm}  \vspace{-0.93006mm} % \vspace{-0.93006mm} % \vspace{-0.93006mm} \vspace{3mm} \vspace{3mm} \vspace{-2mm} \vspace{2mm} \vspace{2mm} \vspace{-2mm} \vspace{-1mm}\vspace{-1mm} \textcolor{blue}{}Global_View_2

\begin{figure*}[htb!]
\centering
\includegraphics[width=\linewidth]{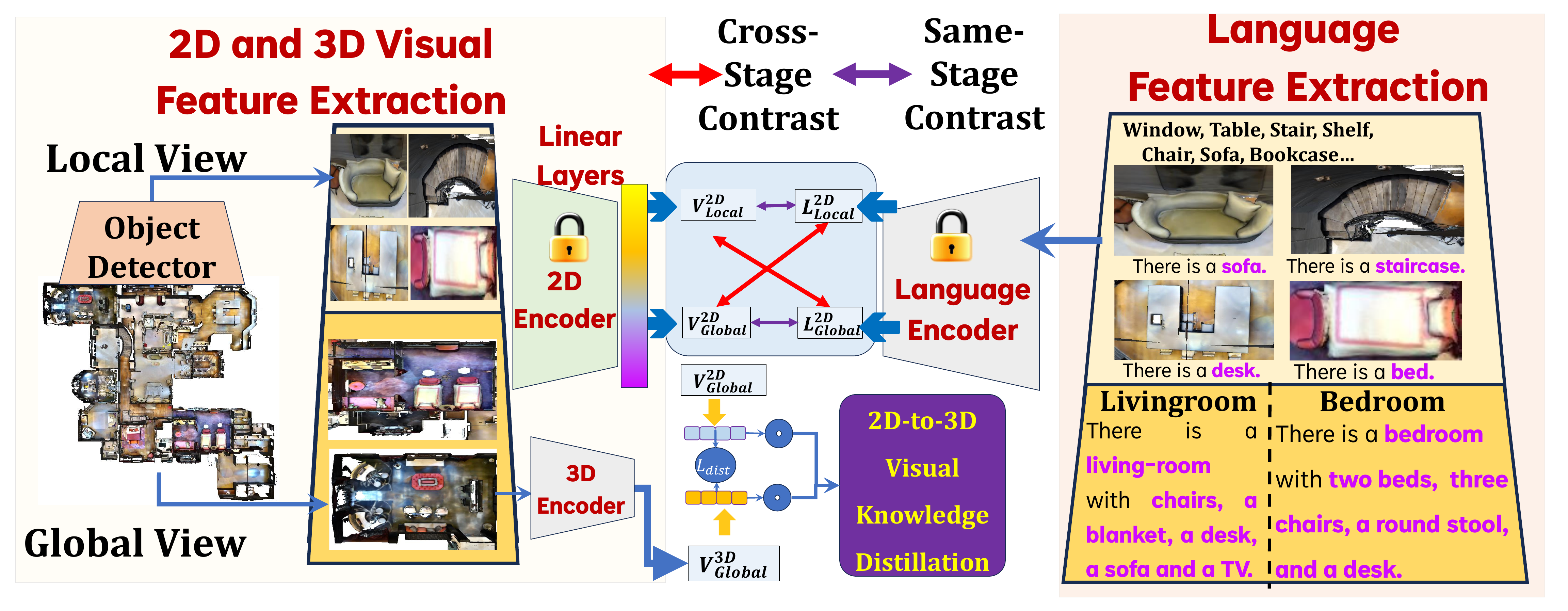}
\caption{\textbf{The pre-training paradigm} of our proposed \textit{WS3D++}. We propose the hierarchical global to local feature alignments to establish the hierarchical vision-language aligned feature representations during the pre-training. This proposed paradigm helps to learn more powerful visual-linguistic aligned feature representation during the pre-training stage. We have further shown the final visualizations and comparisons with CLIP text presentations ranging from both the global view level to the local object category level. The results have further demonstrate the vision-language aligned feature representation for 3D scene parsing.}
\label{fig_frammwork_language}
\vspace{-3mm}
\end{figure*}
% \vspace{-2mm}\vspace{-2mm}\vspace{-3mm}\vspace{1mm}\vspace{3mm}
% % % \vspace{1mm}\vspace{3mm}\vspace{1mm}\vspace{1mm}\vspace{1mm}\vspace{-1mm}\vspace{-2mm}\vspace{-1mm}\vspace{2mm} \vspace{-1mm} \vspace{-2mm} \vspace{-6mm} \vspace{-0.93006mm} \vspace{-3mm}
% \vspace{-2mm} \vspace{-5.8 mm} \vspace{-0.93006mm}
% \vspace{-2mm} \vspace{-5.8 mm}
% \label{Intro} \vspace{-0.510198cm} 
% \vspace{-0.510198cm} \vspace{-1mm} 
% neural  . the  and it , which will be detailed in below. very  or handling s has the capacity to recognize being  of  building quite 
% cannot scale up and the neural 
% tackles the closed-set assumption ion understanding expanding  learned to  rich and 
% These works have demonstrated the ability to further facilitate the extraction of abundant information from the visual representation. different novel model knowledge  {The of when and aligned classes embedding 
\noindent
\textbf{Close-set assumption:} One of the major bottlenecks in scaling up visual perception systems is the poor generalization capacity while encountered with diverse novel semantic classes or severe domain shifts. To endow the model with the capacity for adapting the learned representation and make it conform to different data distributions as well as recognize diverse novel categories, pioneer researches such as CLIP~\cite{radford2021learning}, Flamingo~\cite{alayrac2022flamingo}, and Otter~\cite{li2023otter} have demonstrated the great potentials in learning well-aligned visual linguistic representation from large-scale image-text pairs on the Internet for improving the model generalization capacity. To this end, subsequent approaches have been proposed in establishing abundant vision-language associations for different visual recognition tasks including detection and segmentation using the large-scale vision-language model (VLM)~\cite{luo2023segclip, zhou2022extract, zheng2023regularized}. The paired visual-linguistic feature representation can enable the recognition of a large number of novel objects or concepts with natural language supervision because the visual and the lexical language features are well-matched in their shared semantic feature space. Despite the remarkable performance achieved in developing diverse vision-language foundation models such as SAM~\cite{kirillov2023segment} and SEEM~\cite{zou2023segment} for image-based scene understanding, it remains very difficult for CLIP~\cite{radford2021learning} to benefit downstream 3D scene understanding because it is difficult to raise the feature dimension to 3D and establish explicit correlations or find clear alignments between large-scale scene/object-level 3D point clouds as well as linguistic semantic concepts. Moreover, it is even harder to transfer the informative knowledge to various downstream 3D scene understanding tasks. These limitations severely restrict the scalability of VLM to handle diverse unseen 3D scenes containing diverse novel 3D object categories.

\vspace{+5.9mm}
% novel novel is restricted required scene vision-language models and the new 20 5 2 5 5 2 18 \vspace{-0.10006mm} \vspace{5mm} pre-training paradigm \vspace{-2mm} \vspace{-3mm} % \vspace{-3mm} % \vspace{-3mm} \vspace{-3mm} \vspace{1mm} % \vspace{1mm} \vspace{12mm} 12 \vspace{+0.61mm}\vspace{+1.9mm} \vspace{+5.9mm} PAMI_23_Final_Revise_Global_Local_View

\begin{figure}[tbp]
\centering
\includegraphics[width=\linewidth]{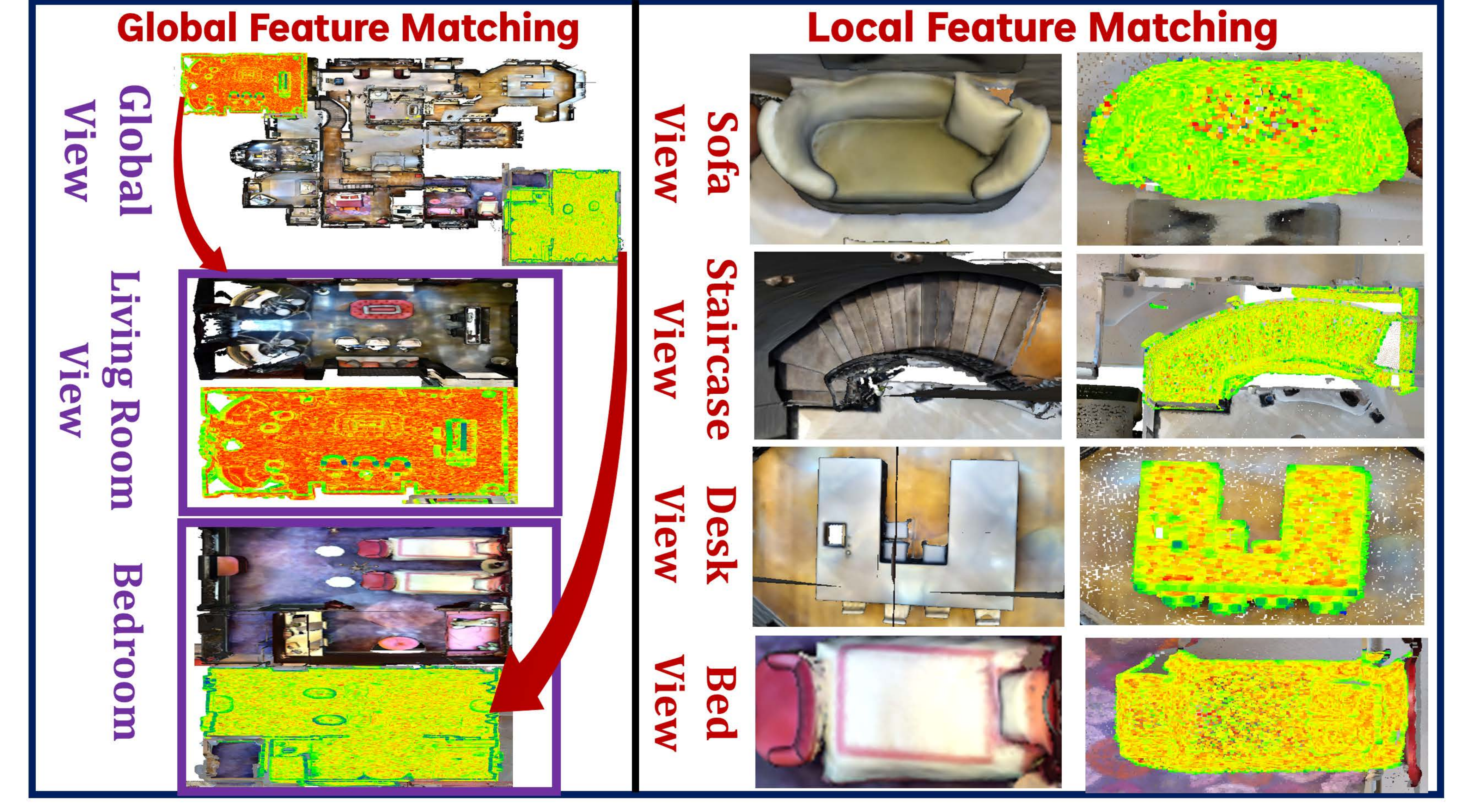}
\caption{\textbf{The feature matching visualization} of our proposed \textit{WS3D++}. We propose hierarchical global to local feature alignments to establish hierarchical vision-language aligned feature representations during pre-training from both the global view level to the local object level. This kind of paradigm helps to learn more powerful visual-linguistic matched representations ranging from both the global view-level to the local object category-level. In the above figure, we have shown the matching at the global view on the left and the matching at local object level on the right. It can be demonstrated that our proposed approach can establish matched feature representation at both the global room feature level and the local object feature level. }
\label{fig_local_feature_matching}
\vspace{-6mm}
% \vspace{-1mm}
\end{figure}
 \vspace{-1mm} \vspace{-3mm}
% \vspace{-3mm}
\begin{figure*}[htb]    
    \centering   \includegraphics[width=\linewidth]{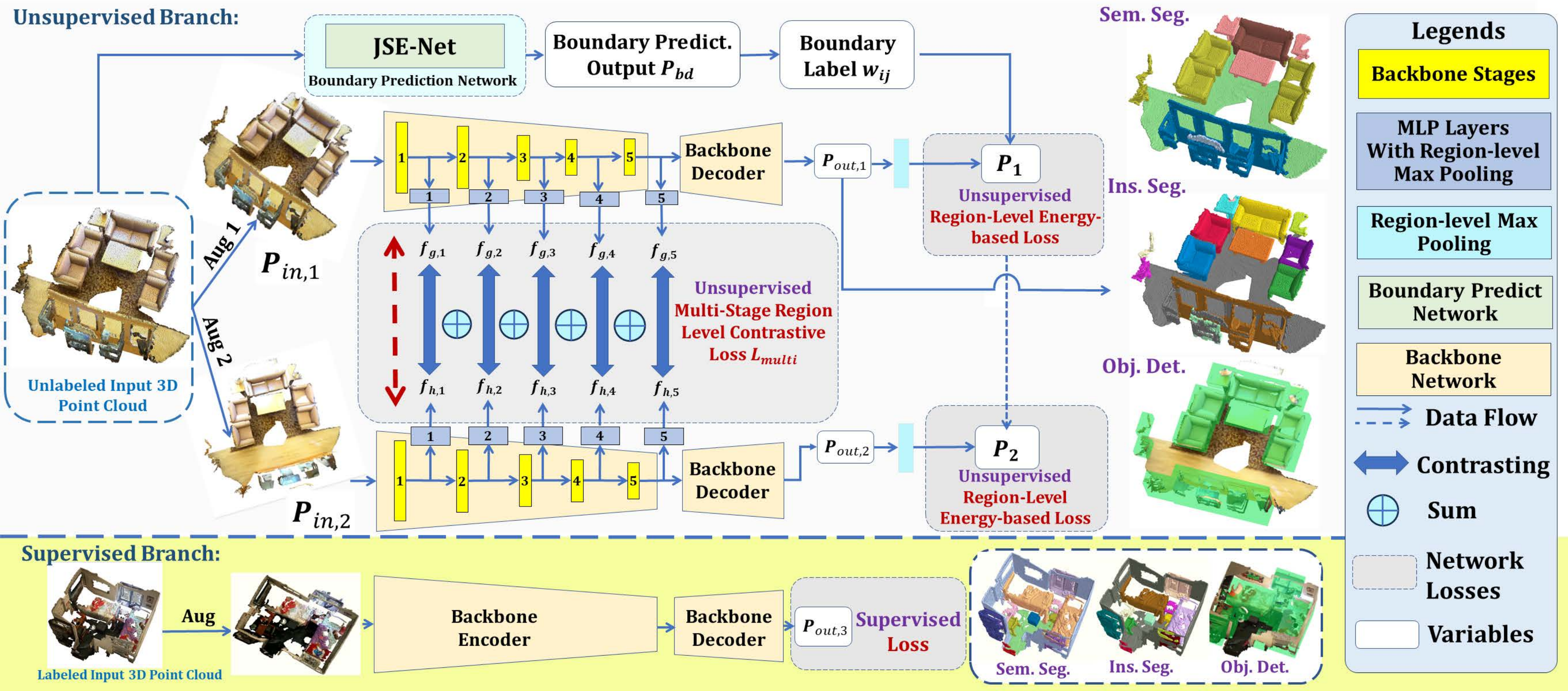}  \caption{The \textbf{fine-tuning paradigm} of our proposed \textit{WS3D++}. \textit{WS3D++}~\cite{liu2022weakly} consists of three proposed modules: 1. The \textbf{unsupervised} region-level energy-based optimization guided by boundary labels; 2. The \textbf{unsupervised} multi-stage region-level contrastive learning with high confidence; 3. The \textbf{supervised} region-level semantic contrastive learning with labeled data. The backbone network adopts encoder-decoder structures. The weights of the backbone network are shared in the supervised and unsupervised branches. Integrated with the proposed pre-training paradigm illustrated in Figure~\ref{fig_frammwork_language}, by our proposed hierarchical feature aligned pre-training and regional fine-tuning, more effective label-efficient learning as well as open-vocabulary learning is realized. }   \label{fig_frammwork}
    \vspace{-1mm}
\end{figure*}

% \textcolor{blue}{
%      \vspace{-1mm}   \vspace{-2mm}\vspace{-0.03006mm}
% kirillov2023segment \vspace{-2mm} \vspace{-1mm} \vspace{-1mm} \vspace{-2mm} \vspace{-2mm} \vspace{-2mm}

% tasks relation  texts images \noindent make when approximately half an hour And many  the  per scene s of the clouds weakly supervised learning (WSL)-based or semi-supervised learning (SSL)-based 3D point cloud understanding Although the image  Weakly supervised focus in recent years applying new 3D labels in the complex for WSL-based semantic understanding, learning requires around reality forty-five of datasets  are
\vspace{-0mm}

\noindent
\textbf{Reliance on large-scale labeled data.}
A critical prerequisite for fully exploiting the capacity of the fully supervised deep learning approaches is the accessibility to large-scale well-annotated high-quality training data. Most point cloud understanding methods rely on heavy annotations~\cite{gong2021omni,behley2019semantickitti, cheng20212}. However, the annotation of large-scale 3D point cloud scenes is rather time-consuming and labor-intensive. For instance, it requires around thirty minutes to label a single scene for ScanNet~\cite{dai2017scannet} or S3DIS~\cite{armeni20163d} with thousands of scenes. Though existing point cloud understanding methods~\cite{gong2021omni, behley2019semantickitti, cheng20212} have achieved decent results on these datasets, it is difficult to directly extend them to novel scenes when the high-quality labeled data is scarce. In the meanwhile, it is often the case that a limited number of scenes can be reconstructed in real applications~\cite{hou2021exploring}. Therefore, developing methods that can be trained with very limited labeled scenes, termed data-efficient 3D scene understanding with limited scene-level annotation, becomes in high demand. Data-efficient semantic and instance segmentation~\cite{shen2023survey, yu2022data, hu2022leveraging} is a vehement research topic for image-level scene understanding. Some simple but successful methods have been proposed, such as contrastive learning~\cite{hu2021region, xie2020pointcontrast} which learns a meaningful and discriminative representation, and conditional random field (CRF)~\cite{obukhov2019gated, chen2017deeplab} for pseudo label propagation. However, there still exist four main challenging unsolved issues while scaling up these approaches to 3D scene understanding. \textit{First}, the widely adopted energy function-based conditional random field segmentation~\cite{chen2017deeplab} relies on handcrafted feature similarities and does not consider explicit boundary information. It attaches equal importance to pixels on semantic boundaries and within same semantic objects, which can cause vague and inaccurate predictions in pixel-level segmentation at object boundaries. And how to leverage boundary information has been explored in 2D but rarely explored in 3D data efficient learning~\cite{rong2023boundary}. \textit{Second}, the computation costs are both very high when applying point-level contrastive learning or point-level energy-based segmentation in a dense point cloud scene for every point pair~\cite{feng2023clustering, liu2023fac}. Furthermore, large-scale point cloud scenes even contain billions of points, making point-level contrastive learning intractable in computational costs.~\textit{Third}, the existing unsupervised contrastive learning-based pre-training for point clouds~\cite{xie2020pointcontrast, hou2021exploring, wiesmann2023kppr, pang2023unsupervised} only considers geometrically registered point/voxel pairs as the positive samples, while it does not explicitly consider explicit regional information, let alone the hierarchical alignments.
\vspace{2mm}

Driven by the above motivations in terms of both generalization capacity and data efficiency, we propose an effective two-stage framework, involving \textbf{unsupervised hierarchical vision-language pre-training} and \textbf{label-efficient fine-tuning} to boost the label/data-efficiency in 3D scene understanding. As shown in Fig.~\ref{Intro}, \textit{in the pre-training stage}, we leverage the rendering techniques to construct well-aligned 2D views for large-scale 3D scenes to establish more accurate coarse-to-fine vision-language associations. Then, we leverage the off-the-shelf object detector and the pre-trained large-scale vision-language model CLIP~\cite{radford2021learning} to construct the \textit{hierarchical feature representations} from both the global scene level to the local object level. We also propose an effective knowledge distillation strategy that distills the informative visual-language-aligned representation of the image encoder in CLIP~\cite{radford2021learning} to the 3D backbone network. As is demonstrated by our extensive experiments, the open-vocabulary scene parsing performance can be also enhanced.
\vspace{-3mm}
% \vspace{-0.2008mm}}
% coarse-grained to fine-grained \vspace{-0.2008mm}}

% view view
% making the alignment 
%Not that our pre-trained database including  ScanNet~\cite{dai2017scannet} for the indoor case and NuScenes~\cite{caesar2020nuscenes} for the outdoor case has clear and accurate correspondence between the 2D images and 3D point clouds, which has made the feature representation alignments among the three diverse modalities including language, image, and point clouds possible.  
% \textit{unsupervised hierarchical vision-language pre-training stage} ies
% representation
% coarse-to-fine multimodal  from the global level to the local level multi-stage unsupervised  3D semantic segmentation and instance segmentation segmentation is achieved final novel  \vspace{3mm}
% \textcolor{blue}{

\textit{During the fine-tuning stage}, as shown in Fig.~\ref{Intro}, we propose a unified \textit{WS3D++} framework that simultaneously solves the 3D scene understanding problem under the data-efficient setting. We first use the over-segmentation~\cite{rusu20113d} to obtain regions and use a boundary prediction network (BPN) as an intermediate tool to obtain boundary region labels. Then, high-confidence boundary region labels serve as the guidance for our proposed region-level energy-based loss. Meanwhile, we propose a region-level confidence-guided contrastive loss to enhance instance discrimination. Specifically, our \textit{WS3D++} includes two innovative designs to address the very challenging label scarcity issues and to enhance performance. Firstly, to encourage latent instance discrimination and to guarantee efficiency, an efficient region-level feature contrastive learning strategy is proposed to guide network training at multiple stages, which realizes the unsupervised instance discrimination. Also, to leverage boundary information as labels for the final semantic divisions, an energy-based loss with guidance from the semantic boundary regions is proposed to take the maximized advantage of the unlabeled data in network training. Combined with supervised loss, the labeled data can also be leveraged to boost the final downstream 3D scene understanding performance. 
\vspace{3mm}
%\vspace{3mm} \vspace{1mm} \vspace{-3mm}Combined with supervised loss, complete 3D scene parsing.  complete 3D scene parsing  of mentioned above 
% Based on 
% The preliminary version of this work is published in the  maximum unsupervised  have utilization our the  published This work \textit{} \vspace{1mm}
% \vspace{1mm}

% \\
 \textit{WS3D++} is a significant extension of the preliminary version of the conference work \textit{WS3D}~\cite{liu2022weakly}, where basic ideas of boundary awareness and contrastive instance discrimination are introduced to tackle the data-efficient 3D scene understanding during fine-tuning stage. In summary, we extensively enriched previous works in the following aspects:
 \vspace{1mm}
% \vspace{-0.0007mm}
% in have the  the Our framework is a generalized framework that combining between diverse modalities of the 
% The new contributions  First Second  weakly supervised  the  stage  during fine-tuning

% In this paper, we propose a generalized framework termed \textit{WS3D++} integrating the vision-language pre-training and the weakly-supervised fine-tuning. Extensive experiments demonstrate that it can be both applied to data-efficient learning and open-vocabulary few-shot learning.  guided the language-aware representations are learned ~\textit{In the meanwhile}, Meanwhile, we visualized the learned representation to demonstrate that discriminative representations are learned within the network.  al .~\textit{Second}, we provided the detailed implementation of our semantic contrastive learning to make them more adaptive to hard samples. Third at both the region level or object level 

% meaningful that make https://sites.google.com/view/zhouyi-joey/team which 

\vspace{-0.001mm}
\textit{First}, we propose a generalized pre-training approach for data-efficient learning, which establishes accurate alignments between language and 3D point cloud in terms of both object-level and scene-level semantics in a hierarchical manner. ~\textit{Second}, we propose to leverage the rendering technique that makes explicit associations between image and point cloud to facilitate 2D-to-3D matching and subsequent language-to-3D matching. ~\textit{Third}, we visualized the language-queried activation maps directly on the 3D scenes, which demonstrates that the proposed approach learns better visual-linguistic alignment between the language descriptions and the visual object-level information.~\textit{Finally}, we evaluate our proposed approach comprehensively in diverse data-efficient and open-world learning settings for both the 3D semantic segmentation and 3D instance segmentation tasks.

%\textcolor[RGB]{0,100, 255}{}  \vspace{-0.19mm} can\vspace{1mm}
% \vspace{-0.28mm} ~\
% detectionsegmentation
% further language prompt
% object-level views
% for providing object-level views
% evaluated our proposed approach in the object detection tasks and demonstrated that the proposed hierarchical representation learning approach during pre-training can provide more accurate instance proposals on the object detection tasks with different settings, which largely benefits both 3D object detection and instance segmentation tested

% ~\textit{Third}, we have extended the proposed approach with more backbone networks, with more experiments on diverse benchmarks and enriched  experimental settings.

% Fourth

% the designed approaches to deal with unseen classes in open-vocabulary settings, which    our proposed methods  designed main  tasks    \vspace{-2.1mm} \vspace{-0.51mm}
% \vspace{0.028mm} \vspace{-0.159mm} \vspace{-1.01mm} 
The contributions of our work are highlighted as follows:
% \vspace{-0.5mm}
% \vspace{-0.02mm} 5

% \vspace{-2mm} label={[\arabic*]} class as well as improve the model generalization capacity to provide high-quality knowledge from open-world categories of vision-language models

% The knowledge distilled from the large-scale vision-language models can mitigate overfitting and render comparatively superior performance in data-efficient learning. At the same time, it can enhance the recognition capacity for novel categories by distilling informative visual-linguistic associations from the visual linguistic models.  tasks
% \vspace{-0.03mm}
\begin{enumerate}[(1)]
% -scale with the aid of a hierarchical manner modality  point cloud the  leveraging effective rendering approaches the the 
% More specifically, we the scene-level point cloud explicitly to  by  ing 0 s design effective approaches 
% \vspace{-0.1mm} an effective approach which \textcolor{blue}{}
\item
During the \textit{pre-training stage}, we first propose an effective design which distills rich knowledge from the large-scale vision-language model into the 3D point cloud modality. Specifically, we propose leveraging rendering to obtain explicit scene-level and object-level 2D-3D feature associations, establishing a more accurate vision-language association hierarchically than the original CLIP encoder. We have demonstrated by extensive experiments that our proposed approach can realize superior compatibility with prevailing weakly supervised approaches. 

% compared which  the  leverage  previous coarse-to-fine  from scene level to region level hierarchical feature representations language-3D most very the provided 3D-2D feature associations unified as a word-to-3D framework for downstream 3D scene parsing. word-to-3D matching unified further \textcolor{blue}{

\item
During the \textit{pre-training stage}, we first propose a global scene-to-sentence matching and then propose a local object-to-word matching approach, respectively, to establish the well-aligned vision-language feature representations at both the scene level and the object level, which largely facilitates the subsequent effective contrastive learning with the mostly matched visual-language contrastive pairs. The proposed designs have both enhanced the data-efficient learning and the knowledge-transfer capacity of the model, as demonstrated by our extensive experiments on both the 3D object detection and the 3D semantic/instance segmentation tasks.

% Therefore, se
% are substantially improved
% accurate 3D shape-language matching  with a high degree of relevance best-matched

% In \item the relative the improved enhanced the - ring in 3D instance loss we design an effective approach to distill knowledge from large-scale vision-language models in a hierarchical manner. It establishes a more accurate coarse-to-fine hierarchical vision-language association.  Therefore, both the data-efficient learning and the knowledge transfer capacity of the model are substantially improved as demonstrated by our extensive experiments on both 3D detection and segmentation tasks.  and  unsupervised  In -level During the  In as  ies al

\item  During the \textit{fine-tuning stage}, we propose a region-aware energy-based optimization approach to achieve the region-level boundary awareness, which utilizes the boundary as additional information to help assist the 3D scene segmentation and understanding. Furthermore, we propose the unsupervised region-level semantic contrastive learning strategy for multi-stage feature discriminations. The energy-based loss and the contrastive loss are jointly optimized for pre-training the backbone network in a complementary manner, which take full advantage of the unlabeled data.

% to the  make full use of the segmentation network the 

%  first \item We propose the first weakly supervised framework that can be simultaneously applied for 3D semantic segmentation and instance segmentation. We conduct lots of experiments on ScanNet and other indoor/outdoor benchmarks such as S3DIS and SemanticKITTI with different annotation ratios. It is demonstrated State-of-the-art performance has been attained.
% \item  First, we propose an unsupervised region-level energy-based loss to achieve region-level boundary awareness, which utilizes boundaries as additional information to assist the 3D scene segmentation.

% \item  We propose the first unsupervised region-level semantic contrastive learning strategy for multi-stage feature discrimination. The energy-based loss and the contrastive loss are jointly optimized for the segmentation network in a complementary manner to make full use of the unlabeled data. are  3D semantic segmentation and instance segmentation first  weakly supervised  downstream various cases 

%We propose a unified framework for data-efficient 3D scene understanding, Combining \textcolor{blue}{} % \vspace{-3mm} \vspace{-1mm} Figure_Seg_SK_PAM_F.png 5

\begin{figure}[t!]
\centering
\includegraphics[width=\linewidth]{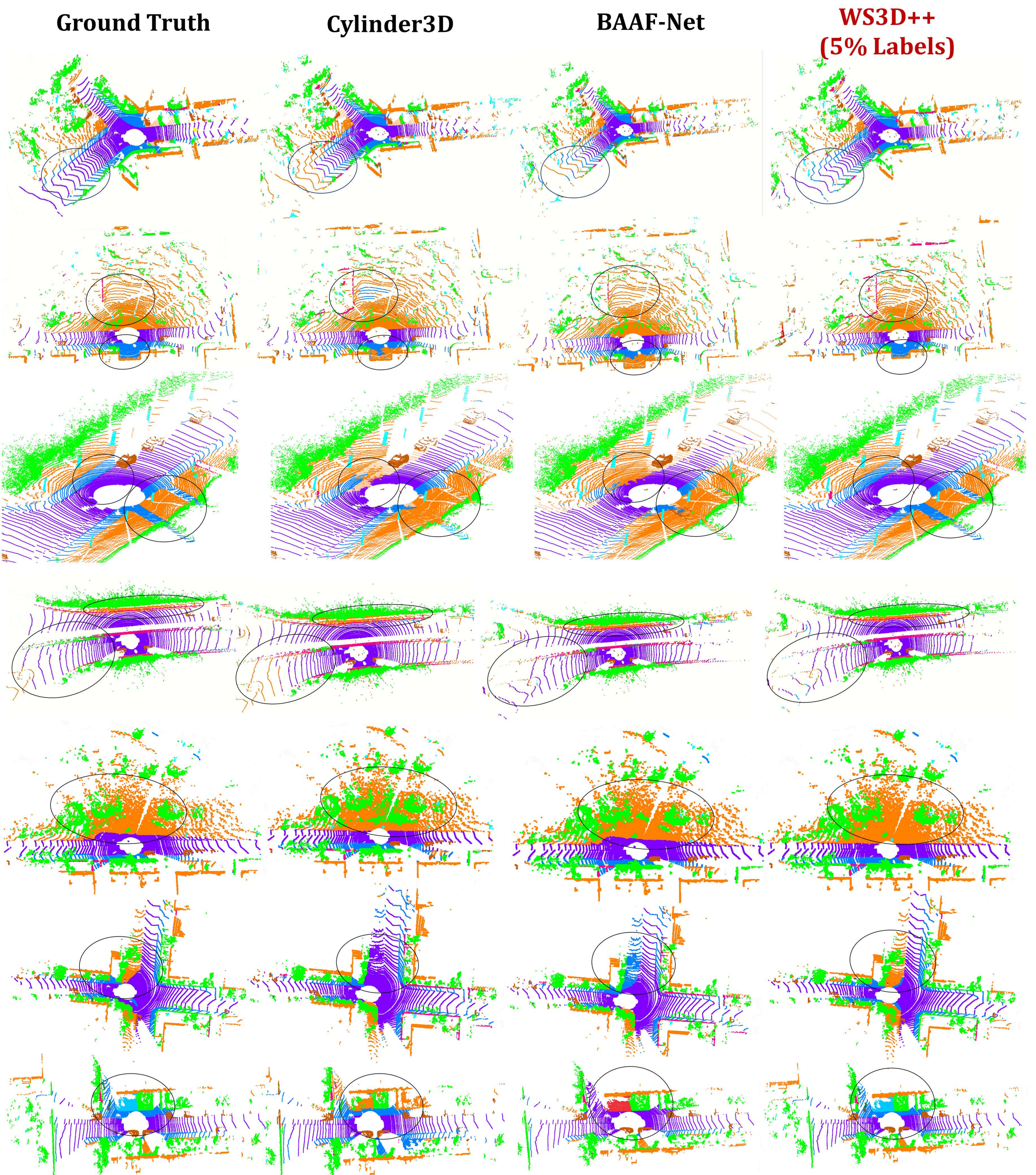}
\caption{Qualitative \textbf{semantic segmentation} results of the proposed \textit{WS3D++} on SemanticKITTI validation set the with the 5\% labeling percentage, compared with the fully supervised state-of-the-art Cylinder3D~\cite{zhu2021cylindrical}, and BAAF-Net~\cite{qiu2021semantic} with the diverse semantics indicated by different colors. The red circles highlight the final performance difference between diverse comparative approaches.}
\label{fig_sem_skitti}
\vspace{-2mm}
\end{figure}

\item Integrating the above two stages as a whole, we propose a unified framework termed \textit{WS3D++}. State-of-the-art performance has been achieved by it with extensive experiments conducted on ScanNet~\cite{dai2017scannet} and other indoor/outdoor benchmarks such as S3DIS~\cite{armeni20163d}, SemanticKITTI~\cite{behley2019semantickitti} and NuScenes~\cite{caesar2020nuscenes} in diverse experimental settings without bells and whistles. Finally, our proposed approach achieves pioneer performance on the very large-scale ScanNet~\cite{dai2017scannet} dataset in diverse downstream tasks of 3D scene understanding, including tasks among 3D semantic segmentation, 3D instance segmentation, and 3D object detection\footnote{\href{http://kaldir.vc.in.tum.de/scannet_benchmark/data_efficient/lr_semantic_instance_3d}{http://kaldir.vc.in.tum.de/scannet\_benchmark/data\_efficient/}}. 
\vspace{-3.93388 mm}
% \vspace{-1mm} 3
% semantic segmentation, ~ pioneering pioneer \vspace{-4.8mm} 6  instance segmentation, and 3D o 1 5

% In the limited reconstruction case, which is a common case in robotic applications where a limited number of scenes are available, 1
\end{enumerate}

% termed 3D \textit{WS3D++}, 
% We ascribe the data ge significantly  ranks 1st it can be seen that combined the endeavor to design better deep neural network structures Therefore,  And o . our contributions any designs capacity of designed downstream evaluates \textit{WS3D++} proposed  framework . learning s an 1 
% \vspace{-0.01888800mm}

To the best of our knowledge, this is the pioneer work which comprehensively evaluates across diverse 3D label-efficient scene understanding downstream tasks with our proposed 3D open-vocabulary recognition approach termed \textit{WS3D++}. Our endeavor is orthogonal to the 3D backbone network designs and thus can be seamlessly integrated with the prevailing 3D point cloud detection or segmentation models.  Our comprehensive results provide solid baselines for future researches in the data-efficient 3D scene understanding. 

% \vspace{-0.1mm} 8 \vspace{+1.83388 mm}\vspace{-3.93388 mm}}
% % \vspace{-1.23388 mm}\vspace{-0.35100501mm} \vspace{-0.3mm} \vspace{-0.1mm} \vspace{-0.1mm}
% \vspace{-0.213388 mm}
% 5 0  45 3 2 3 2 5 1  \vspace{-2.68mm} Finally, w have provided \vspace{-0.0006mm} 3 -2.38mm across diverse current mainstream for  1.18mm 1 \vspace{-2.88mm} \vspace{-2.68mm} \vspace{-3.98mm} 1 5 3 3  
% which \vspace{-0.0mm}
% WSL-based It has been demonstrated by extensive experiments that 
% architectures LiDAR segmentation methods Improving Underwater ROV Inspections via Learnable Physical Model‑Guided Unsupervised
% Image Enhancement 7.18mm \vspace{-0.03388 mm} \vspace{-3mm} \vspace{-3mm} \vspace{-3mm} \vspace{-3mm} \vspace{-3mm} t! \vspace{-3mm} \vspace{-2mm} \vspace{2mm} \vspace{-5mm} \vspace{-5mm} \vspace{-3mm}

\begin{figure}[t!]
% h\setlength{\abovecaptionskip}{-0.0101cm}
% \setlength{\belowcaptionskip}{-0.00101cm} % \vspace{-5.88mm} \vspace{-1.88mm} unitive combinative can be our 2 \vspace{-6mm}
\centering
\includegraphics[width=\linewidth]{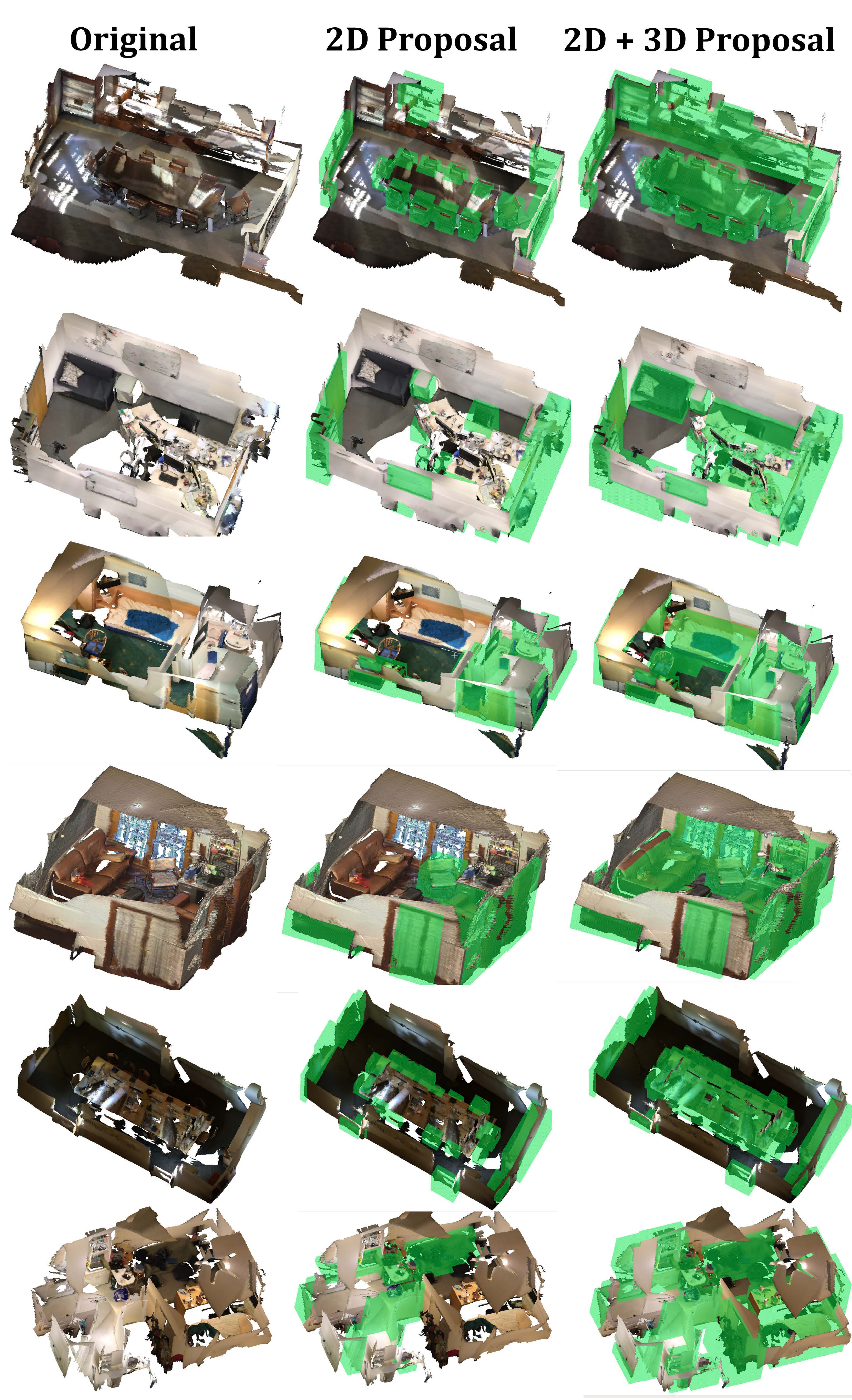}
\caption{The captured 2D and 3D region proposals. It is demonstrated qualitatively clearly that more precise object proposals are captured by proposed united 2D/3D proposal generation approach. It can be demonstrated that clear superior regional proposal generation performance can be well guaranteed. }
\label{fig_render_pro}
\vspace{-6mm}
\end{figure}

\begin{figure}[t]
% % \vspace{-5mm}\vspace{-1mm}regional feature \vspace{-1.88mm}\setlength{\abovecaptionskip}{-0.12cm} \vspace{-0.1mm}
% \setlength{\belowcaptionskip}{-0.21cm} \vspace{-1.88mm} differences in detection \vspace{-6mm} \vspace{-3mm}
\centering
\includegraphics[width=\linewidth]{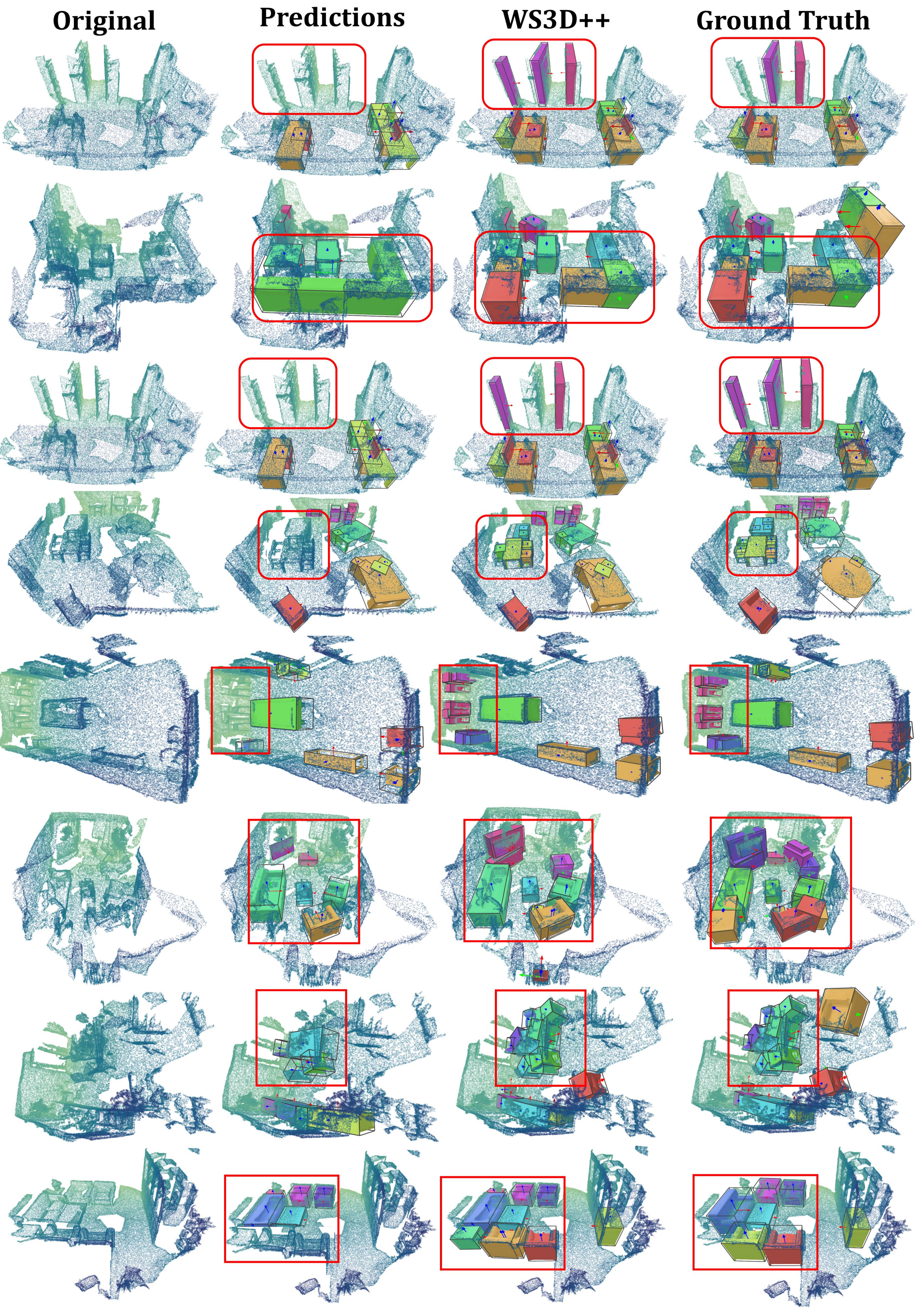}
\caption{The final ScanNet~\cite{dai2017scannet} object detection results and performance. The comparative differences in detection predictions are highlighted in the rectangles. }
\label{fig_scannet_det}
\vspace{-3mm}
\end{figure}

\section{Related Work}

% \vspace{-0.03388 mm} \vspace{-1mm}
% \vspace{-0.38mm} \textbf{} ,tang2020searching ,ye2020hvnet , vu2022softgroup
\vspace{-0.06mm}
\textbf{Learning-based Point Cloud Understanding.} Deep-network-based approaches are widely adopted for point cloud understanding. Fully supervised approaches can be roughly categorized into voxel-based~\cite{que2021voxelcontext,choy20194d, noh2021hvpr, yang2022towards, vu2022softgroup++}, projection-based~\cite{wu2019squeezesegv2, xu2020squeezesegv3, feng2018gvcnn, kundu2020virtual,li2020end,gojcic2020learning}, and point-based approaches~\cite{landrieu2019point,zhang2019shellnet,yan2020pointasnl,liu2019densepoint,yin2021center,ao2021spinnet,fan2020pstnet,lei2020spherical, liu2019point, huang2021predator, liu2021fg}. The voxel-based approaches~\cite{graham20183d, chen2023largekernel3d, liu2022spatial} which are built upon SparseConv~\cite{graham20183d} and voxelize the point cloud for efficient processing have achieved remarkable performance in 3D scene parsing. Therefore, we use SparseConv~\cite{graham20183d} as our backbone architecture for downstream semantic understanding tasks because of its high performance in inferring 3D semantics. \\

\vspace{-2.58mm}
% segmentation and detection network  The typical point-based method is the super point-graph \cite{landrieu2019point} proposing graph-based deep metric learning for point cloud over-segmentation, which has inspired our work. Different from them, we use the over-segmentation result as the intermediate tool to obtain the boundary region labels.
 % it
% both LiDAR-based scene 
% semantic segmentation The typical voxel-based method is the Sparseconv \cite{graham20183d}.
 % the Point Clouds Understanding. ~\cite{xie2020pointcontrast}  on the target dataset fully labeled ground truth or points  the high-level supervised point cloud semantic segmentation~\cite{eckart2021self} contrastive s
\vspace{-1mm}
% e ive {-0.01806cm} 2 {-0.01206cm} 12

\noindent
\textbf{Pre-training for 3D Representation Learning.} Many recent works propose to pre-train networks on source datasets with auxiliary tasks such as low-level point cloud geometric registration~\cite{xie2020pointcontrast}, 3D local structural prediction~\cite{thabet2020self}, the completion of the occluded point clouds~\cite{wang2021unsupervised}, and the foreground-background feature discrimination~\cite{liu2023fac}, with effective learning strategies such as contrastive learning~\cite{xie2020pointcontrast} and masked generative modelling~\cite{eckart2021self, wu2023masked}. Then they finetune the weights of the trained networks for the downstream target tasks to boost their performances. However, several major challenges still exist. \textit{First}, the above pre-training approaches all rely on the closed-set assumption, which means that the model can barely be transferred to recognize novel categories that do not appear within the training data. \textit{Second}, the above methods require accessibility to the well-registered augmented point cloud~\cite{xie2020pointcontrast, hou2021exploring, wiesmann2023kppr, pang2023unsupervised} to construct the pre-training contrast views, which are very hard to obtain for large-scale 3D scenes. \textit{Third}, a large number of computational power is required in the pre-training stage. Therefore, the designed pre-training approaches need to be very simple and lightweight, thus will make it easier to directly transfer the model to large-scale point clouds.

\vspace{-0.298mm}
%Otherwise, they can not be easily transferred to large-scale point clouds.  3D understanding  in the pre-training  can be  several  The remarkable performance Different from them, we explore an effective 3D recognition paradigm using the prevailing pre-training and fine-tuning pipeline. Flamingo \textcolor{blue}{

Recently, with the development of large-scale vision-language models such as CLIP~\cite{radford2021learning} and Flamingo~\cite{alayrac2022flamingo}, we can largely benefit the recognition capacity of 3D scene understanding models by distilling informative knowledge from large-scale vision-language models. For example, some pioneer works such as PointCLIP~\cite{zhang2022pointclip, zhu2023pointclip} and ULIP~\cite{xue2023ulip} successfully transfer the knowledge from vision-language models to boost the downstream 3D shape classification tasks. The 3D CAD shapes are transferred to multi-view depth maps, thus they can be fed into the CLIP visual encoder and used the image representations paired with the corresponding point cloud as a bridge to obtain the correlations between the 3D and textual features. Moving beyond object-level recognition tasks, pioneer works also explore how to establish alignment among images, language, and 3D point cloud scenes for the task of open-vocabulary 3D scene understanding~\cite{ding2023pla,ding2023lowis3d,yang2023regionplc}, which target localizing, detecting, and segmenting novel object categories that do not exist in the annotation. Compared with them, our proposed simple but effective framework can be both applicable to data-efficient learning and open-vocabulary scene understanding. \\
% \vspace{1mm}}\vspace{-1.58mm}
\vspace{-2.58mm}

% understanding applied 0 7 0 0 3  \vspace{-2.58mm}\vspace{-5.8mm} \vspace{-5.8mm} PAMI_23_Final.png

% \setlength{\abovecaptionskip}{-0.01066cm}
% \setlength{\belowcaptionskip}{-0.01506cm} \vspace{-1.2006mm} \vspace{-0.1006mm} \vspace{-0.2006mm} 3_Final PAMI_23_Final PAMI_23_Final.png \includegraphics[width=\linewidth] PAMI_23_Final_Revise.png Global_View kind of

% Global_View.png h fig/PAMI_23_Final_Revise_Global_Local_View.png

%  %  % \vspace{-0.60006mm}\vspace{-2.60006mm}\vspace{-2.60006mm} 5\vspace{+3.20006mm}\vspace{-3.2006mm} % \vspace{+0.10006mm}\vspace{-0.10006mm}2 3 Although remarkable performance achieved, the tested scenarios are restricted to the specific indoor dataset. with open-vocabulary recognition capacity  current  less labeled data and yang2023regionplc has been learning \vspace{-0.2006mm}
 % Also, we aim at providing a generalized solution that can be applicable for both indoor and outdoor point cloud scenes and a wide range of 3D scene understanding tasks including both detection and segmentation. In this work 6 \vspace{-0.2006mm} \vspace{-0.2006mm} \vspace{-0.80006mm}
\vspace{-0.9mm}

\noindent
\textbf{Label/Data-Efficient Learning for 3D.} Recent studies have produced many elaborately designed backbone networks for 3D semantic/instance segmentation~\cite{hu2020randla, qiu2021semantic,li2020end, jiang2020pointgroup, ding2022doda}, as well as for 3D object detection~\cite{shi2020pv, shi2020points, shi2023pv}. However, they rely on full supervision. Directly applying current SOTAs (State-of-the-art) methods for training will result in a great decrease in performance~\cite{hu2021sqn} for WSL, if the percentage of labeled data drops to a certain value, e.g., less than 30\%. Recently, many works have started to focus on point cloud semantic segmentation with partially labeled data. Wang et al.~\cite{wang2020weakly} choose to transform point clouds to images, but pixel-level semantic segmentation labels are required for network training. Sub-cloud annotations~\cite{wei2020multi} require extra labor to separate the sub-clouds and to label points within the sub-clouds. Liu~\cite{liu2022rm3d} proposed a robust data-efficient 3D scene parsing framework. It leverages the complementary merits of the superior generalization capacity of the traditional 3D descriptors and the strong feature description capacity learned 3D descriptors to learn very robust local features. Then using the descriptor guided learned region merging~\cite{zhang2023growsp}, superior performance can be achieved on downstream tasks. Liu et al.~\cite{liu2021one, liu2023one} proposed self-training techniques to tackle scene understanding in weak supervision, which design a two-stage training scheme to produce iteratively optimized pseudo labels from weak labels during training. Despite satisfactory results, these approaches still have not learned generalized representation applicable to diverse tasks. Xu et al.~\cite{xu2020weakly} adopt a weakly-supervised training strategy, which combines training with coarse-grained information and partial points using 10\% labels. However, their tested cases are limited to object part segmentation, and it is difficult to uniformly choose points to label. The convex decomposition~\cite{gadelha2020label} is conducted in an approximate manner to perform 3D scene parsing on the object parts. More approaches~\cite{li2022coarse3d} have been proposed recently, which utilize class prototypes and masked point cloud modeling~\cite{liu2023cpcm, xu2023mm, wu2023masked} to learn informative representations for downstream 3D scene understanding. Conceptfusion~\cite{jatavallabhula2023conceptfusion} has also been proposed to alleviate the labelling burden via rendering with the proposed CLIP-driven queryable 3D point cloud maps. To sum up, although approaches have been proposed to alleviate the data efficiency problem, the models for weakly supervised learning lack the capacity to recognize novel categories beyond the labeled training set. Our framework tackles open-set and data-efficient learning problems and is widely applicable to diverse 3D scene understanding downstream tasks.

\begin{table*}[t]
    \centering \caption{Comparisons of the open-vocabulary learning performance on ScanNet. It can be demonstrated that our proposed approach provides very superior open-world recognition performance compared with the diverse SOTAs. The results are given as hIoU / {mIoU}$_{\mathcal{B}}$ / {mIoU}$_{\mathcal{N}}$, respectively. We have compared intensively with diverse labeled ratios. We apply panoptic quality mean intersection over union (mIoU), which can be divided into segmentation quality and recognition quality as metrics for the instance segmentation task. These evaluation metrics are computed on base (B) and novel (N) categories, with the superscripts of B and N (e.g. {mIoU}$_{\mathcal{B}}$ / {mIoU}$_{\mathcal{N}}$), respectively. Furthermore, we have also utilized the harmonic metric such as harmonic IoU (hIoU) as the major indicators for open-world tasks following popular zero-shot learning work to consider the partition between base and novel classes.  }  

    \renewcommand\arraystretch{1.32}
    \label{table_open_world_scannet}    \resizebox{\linewidth}{!}{
\begin{tabular}{c|l|l|l|l|l|l|l}
    \toprule
    \multirow{2}{*}{Datasets} & \multirow{2}{*}{Models} & \multicolumn{3}{c|}{80\%} & \multicolumn{3}{c}{100\%} \\
    \cline{3-8}
    && \multicolumn{1}{c|}{B15/N4} &\multicolumn{1}{c|}{B12/N7 } &\multicolumn{1}{c|}{ B10/N9} & \multicolumn{1}{c|}{B15/N4} &\multicolumn{1}{c|}{B12/N7 } &\multicolumn{1}{c}{ B10/N9 } \cr
    \cline{3-8}
    %  [-1em] &
    % &  Induct. & Transduct.&Induct. & Transduct. & Induct. & Transduct. \\ &    
  % Merely the pre-training stage & 42.9 / 69.9 / 33.8 &  37.8 / 66.6 / 28.8 &  20.1 / 67.7 / 15.1 &  43.8 / 72.8 / 39.6 & 43.8 / 68.9 / 32.8 & 24.5 / 68.7 / 27.1 \\  

  \hline
  \multirow{12}{*}{ScanNet~\cite{dai2017scannet}} 

  % Baseline & 27.2 / 56.8 / 13.8 & 23.6 / 36.6 / 16.9 & 13.9 / 63.9 / 6.8 & 36.9 / 72.7 / 22.7 & 25.5 / 45.3 / 38.6 & 24.5 / 66.7 / 27.1 \\
  % &    Merely the pre-training stage & 42.9 / 69.9 / 33.8 &  37.8 / 66.6 / 28.8 &  20.1 / 67.7 / 15.1 &  43.8 / 72.8 / 39.6 & 43.8 / 68.9 / 32.8 & 24.5 / 68.7 / 27.1 \\

 & Baseline & 39.7 / 48.8 / 38.9  & 37.9 / 44.9 / 42.6  & 31.9 / 38.8 / 33.9  & 43.6 / 50.7 / 39.6  & 38.8 / 48.9 / 45.8 & 33.9 / 42.9 / 38.8 \\
  
 & Merely the Pre-training Stage & 55.9 / 52.9 / 48.8  & 46.9 / 48.8 / 46.9  & 43.9 / 46.7 / 43.8  & 69.6 / 67.8 / 59.8  & 55.8 / 50.9 / 48.8 & 53.6 / 47.9 / 46.8 \\
  & 3DGenZ~\cite{cheraghian2020transductive} & 32.6 / 76.8 / 26.6 & 25.9 / 46.5 / 26.3 & 28.3 / 63.6 / 18.6 & 42.9 / 76.9 / 39.6 & 32.9 / 56.8 / 33.9 & 27.1 / 69.8 / 32.8 \\
  & 3DGenZ~\cite{cheraghian2020transductive} (Pr) & 42.8 / 78.8 / 42.9 & 39.6 / 68.3 / 51.8 & 27.6 / 79.5 / 18.9 & 46.8 / 88.6 / 48.6 & 43.8 / 78.8 / 49.9 & 37.9 / 83.7 / 38.9 \\
    &3DTZSL \cite{michele2021generative} & 17.6 / 45.8 / 6.1 & 23.8 / 36.6 / 12.8 & 13.8 / 67.5 / 15.7 & 42.9 / 49.6 / 16.7 & 27.8 / 38.9 / 15.9 & 16.9 / 69.8 / 18.8 \\
    &3DTZSL \cite{michele2021generative} (Pr) & 49.6 / 58.7 / 48.9  & 38.8 / 50.6 / 38.6 &  33.9 / 69.8 / 33.9 &  59.6 / 69.8 / 59.6 & 48.9 / 59.6 / 48.7  & 38.6 / 73.9 / 39.6 \\
    &LSeg3D \cite{wang20213dioumatch} & 0.0 / 64.4 / 0.0 & 26.9 / 55.7 / 22.8 & 11.8 / 77.6 / 23.9 & 33.6 / 76.8 / 33.8  & 28.8 / 72.6 / 29.2 &  26.6 / 68.5 / 26.9 \\
    &LSeg3D \cite{wang20213dioumatch} (Pr) & 38.9 / 78.8 / 29.9 & 32.8 / 73.6 / 26.8 & 28.6 / 66.8 / 22.8 & 49.6 / 82.7 / 46.9 & 46.9 / 76.8 / 43.2 &  38.9 / 70.3 / 39.6  \\
    &PLA without caption \cite{ding2023pla}& 39.7 / 73.3 / 28.0 & 31.5 / 72.3 / 25.6 & 28.7 / 68.6 / 23.5 & 45.7 / 72.9 / 36.9 &  31.9 / 77.8 / 26.8 & 29.8 / 79.8 / 28.3 \\

    &PLA without caption~\cite{ding2023pla} (Pr) & 53.9 / 82.6 / 45.8 & 48.6 / 80.6 / 40.8 & 42.6 / 75.5 / 37.8 & 56.7 / 84.3 / 52.9 & 53.9 / 83.8 / 45.3 &  52.1 / 80.8 / 45.1 \\
      
    &PLA \cite{ding2023pla} & 69.2 / 70.6 / 66.9 & 60.9 / 68.9 / 59.9 & 
    58.2 / 78.6 / 73.8 & 69.3 / 73.3 / 72.2 & 62.7 / 73.6 / 61.3 & 65.6 / 81.8 / 76.8 \\

    &PLA (Pr)~\cite{ding2023pla} & 79.9 / 82.9 / 81.9 & 78.5 / 80.6 / 68.9 & 
    65.9 / 77.8 / 57.8 & 82.5 / 88.2 / 84.9 &  80.8 / 85.7 / 73.9 &  76.8 / 85.3 / 63.9 \\

    % & \textit{WS3D-Open} (Ours)  &  68.3 / 70.6 / 67.1 &  64.2 / 71.4 / 59.9  & 59.7 / 77.9 / 51.6 & 69.6 / 71.6 / 75.5  &  67.9 / 73.5 / 72.7 &  65.6 / 81.5 / 62.9 \\

    % & \textit{WS3D-Open} (Ours) (Pr) \cellcolor{orange!37}  & 69.6 / 75.8 / 70.9 \cellcolor{orange!37}  &  \cellcolor{orange!37} 67.8 / 78.9 / 67.8 &  \cellcolor{orange!37}  
    % 66.9 / 86.5 / 65.7 & \cellcolor{orange!37}  72.3 / 80.6 / 76.6  &  \cellcolor{orange!37} 71.9 / 89.8 / 78.9 &  \cellcolor{orange!37}  68.8 / 87.9 / 68.8 \\

   &  \textit{WS3D++} (Ours)  &  72.3 / 69.1 / 73.3   &    68.6 / 65.6 / 68.6  &   73.2 / 71.9 / 71.6 & 77.9 / 76.6 / 78.6 & 73.8 / 72.8 / 75.8 &  71.9 / 72.8 / 73.9  \\

   & \cellcolor{orange!37} \textit{WS3D++} (Ours) (Pr) & \cellcolor{orange!37} 75.8 / 73.8 / 77.5 & \cellcolor{orange!37} 78.6 / 75.3 / 78.9 & \cellcolor{orange!37} 71.8 / 85.6 / 79.5 & \cellcolor{orange!37} 86.7 / 83.7 / 85.8 & \cellcolor{orange!37} 82.7 / 80.9 / 83.6 &  \cellcolor{orange!37} 78.9 / 88.9 / 83.8
 %    & Fully-supervised & 74.5 / 68.4 / 79.1 & 73.6/ 
 % 72.0 / 72.8 &  69.9 / 75.8 / 64.9 & 76.7 / 69.8 / 17.9  &  19.9 / 68.5 / 69.7 &  69.2 / 68.5 / 69.2 \\

 %    & Fully-supervised (Pr) \cellcolor{orange!37} & \cellcolor{orange!37} 88.6 / 87.6 / 85.6 &  \cellcolor{orange!37} 87.2 / 83.2 / 83.6 & \cellcolor{orange!37} 83.9 / 88.9 / 77.9 & \cellcolor{orange!37}  83.9 / 83.8 / 89.3 & \cellcolor{orange!37}  83.9 / 89.5 / 89.8 & \cellcolor{orange!37}  86.8 / 88.6 / 82.6 
   \cr
    \cline{3-5}
        \hline

% B12/N3  Merely the pre-training stage  &  23.6 / 69.7 / 19.6  &  23.1 / 35.8 / 22.8 &  14.8 / 65.7 / 11.6 &  23.9 / 66.8 / 19.9 &  26.6 / 43.8 / 29.8 &  23.7 / 73.6 / 12.8  \\  
    % & 

     \hline  
      \multirow{2}{*}{Datasets} & \multirow{2}{*}{Models} & \multicolumn{3}{c|}{10\%} & \multicolumn{3}{c}{50\%} \\
     \cline{3-8} 
     && \multicolumn{1}{c|}{B15/N4} &\multicolumn{1}{c|}{B12/N7} &\multicolumn{1}{c|}{B10/N9} & \multicolumn{1}{c|}{B15/N4} &\multicolumn{1}{c|}{B12/N7} & \multicolumn{1}{c}{B10/N9} 
    \cr
    \cline{3-5}
        \hline
      
    \multirow{12}{*}{ScanNet~\cite{dai2017scannet}} & 

    Baseline & 27.9 / 18.2 / 21.9 & 23.0 / 33.6 / 31.2 &  19.6 / 28.6 / 27.6 & 35.1 / 20.6 / 36.6 & 31.7 / 41.5 / 38.2 & 28.5 / 32.3 / 28.2 \\
    & 
    
    Merely the Pre-training Stage & 35.8 / 58.3 / 36.8  & 33.8 / 37.9 / 31.8 & 30.8 / 70.3 / 25.6 & 37.2 / 63.2 / 39.7 & 48.9 / 53.8 / 49.9 & 49.6 / 86.8 / 49.8 \\
    & 
    3DGenZ~\cite{cheraghian2020transductive} & 17.3 / 52.3 / 9.7  & 15.3 / 28.8 / 12.5 & 11.2 / 62.5 / 5.2 & 18.6 / 52.8 / 11.6 & 17.8 / 33.6 / 12.9 & 11.8 / 72.3 / 6.3 \\
    & 3DGenZ~\cite{cheraghian2020transductive} (Pr) & 28.5 / 72.9 / 28.2 & 23.6 / 38.9 / 23.6 & 15.6 / 68.9 / 12.9 & 29.7 / 75.6 / 29.7 & 29.5 / 45.3 / 45.7 & 23.3 / 43.9 / 11.9 \\
    &3DTZSL \cite{michele2021generative} & 8.9 / 33.9 / 6.1 & 3.3 / 35.8 / 1.8 & 7.3 / 47.3 / 2.8 & 16.2 / 32.5 / 23.2 & 21.9 / 31.3 / 6.8 & 9.9 / 63.5 / 8.8 \\
    &3DTZSL \cite{michele2021generative} (Pr) & 26.8 / 43.8 / 28.7  &  18.6 / 52.6 / 23.2 &  16.8 / 49.5 / 19.2 &  31.3 / 49.3 / 28.7 & 28.6 / 43.9 / 27.3  & 15.3 / 66.9 / 21.7 \\
    &LSeg3D \cite{wang20213dioumatch} & 0.0 / 55.6 / 0.0 & 0.6 / 43.6 / 0.3 & 1.2 / 56.1 / 0.8 & 19.9 / 65.3 / 13.6  &  12.9 / 56.8 / 8.7 &  9.9 / 58.5 / 8.7 \\
    &LSeg3D \cite{wang20213dioumatch} (Pr) & 28.8 / 65.3 / 23.8 & 26.7 / 62.8 / 16.7 & 23.7 / 62.8 / 22.9 & 32.8 / 76.7 / 28.7 &  28.9 / 73.9 / 26.9 & 26.8 / 73.2 / 25.6  \\
    &PLA without caption \cite{ding2023pla} & 35.3 / 53.2 / 18.2 & 26.7 / 52.1 / 9.9 & 19.6 / 56.7 / 9.7 & 36.5 / 55.9 / 22.8 &  28.2 / 58.7 / 18.8 &  25.8 / 62.1 / 15.2 \\

    &PLA without caption~\cite{ding2023pla} (Pr) & 39.3 / 73.3 / 35.7 & 28.6 / 67.8 / 29.6 & 24.5 / 62.6 / 26.8 & 42.9 / 75.6 / 41.8 &  39.3 / 73.6 / 36.9 &  35.6 / 72.3 / 33.8 \\
      
    &PLA \cite{ding2023pla} & 51.3 / 58.8 / 48.6 & 49.3 / 55.9 / 44.9 & 44.7 / 52.2 / 40.8 &  65.7 / 72.8 / 65.7 &  58.8 / 66.5 / 49.8  & 53.6 / 63.5 / 46.8 \\

    &PLA (Pr)~\cite{ding2023pla} & 61.8 / 67.6 / 73.9 & 58.7 / 59.3 / 56.8 & 
    68.8 / 82.9 / 47.6 & 72.5 / 82.8 / 79.9 & 71.8 / 81.9 / 61.8 & 70.3 / 83.8 / 49.8 \\
     % & RegionPLC \cite{yang2023regionplc}& 69.9/68.4/71.5 & 65.1/69.6/61.1 & 58.8/76.6/47.7 \\ 1.18  \cellcolor[RGB]{232,232,232} \cellcolor[RGB]{232,232,232} \cellcolor[RGB]{232,232,232} \cellcolor[RGB]{232,232,232}  \cellcolor[RGB]{232,232,232} \cellcolor[RGB]{232,232,232} \cellcolor[RGB]{232,232,232} \cellcolor[RGB]{232,232,232}  \cellcolor[RGB]{232,232,232} \cellcolor[RGB]{232,232,232} \cellcolor[RGB]{232,232,232} \cellcolor[RGB]{232,232,232}  \cellcolor[RGB]{232,232,232} \cellcolor[RGB]{232,232,232} \cellcolor[RGB]{232,232,232}   \cellcolor{orange!37} \cellcolor{orange!37}
     % &  \textit{WS3D} (Ours) & 66.1/69.1/65.5 &  63.3/70.3/58.8 &  \cellcolor[RGB]{232,232,232}
     %  59.3/77.5/51.6 \\     \multirow{15}{*}{ScanNet~\cite{dai2017scannet}} & 3DGenZ~\cite{cheraghian2020transductive} & 20.6/56.0/12.6& 19.8/35.5/13.3 & 12.0/63.6/6.6\\

     % & \textit{WS3D} (Ours) (Pr) \cellcolor{orange!37} & \cellcolor{orange!37} 68.5/77.8/78.6 &  \cellcolor{orange!37} 66.2/75.7/65.7 &  \cellcolor{orange!37}
     % 67.8/86.9/67.6 \\ Few-shot settings Few-shot settings
      
    % & \textit{WS3D-Open} (Ours)  &  55.3 / 56.6 / 53.1 &  54.2 / 63.6 / 53.9  & 59.7 / 77.9 / 51.6 &  63.6 / 62.9 / 65.9  &  65.6 / 73.5 / 67.9 &  55.6 / 66.8 / 67.7 \\

    % &  \textit{WS3D-Open} (Ours) (Pr) \cellcolor{orange!37}  & 66.9 / 63.9 / 59.8 \cellcolor{orange!37} &  \cellcolor{orange!37} 59.6 / 58.2 / 55.7 &  \cellcolor{orange!37}  
    % 53.7 / 55.2 / 53.7 & \cellcolor{orange!37}  68.7 / 67.8 / 66.9 &  \cellcolor{orange!37}  62.7 / 71.2 / 65.5 &  \cellcolor{orange!37} 65.8 / 63.9 / 56.8 \\

   &  \textit{WS3D++} (Ours)   &  65.9 / 75.2 / 68.3  &  61.3 / 65.6 / 62.7  &  58.6 / 60.9 / 56.8  & 71.8 / 76.8 / 81.6  & 65.8 / 69.5 / 68.9 &  62.8 / 67.8 / 68.2 \\

   & \cellcolor{orange!37} \textit{WS3D++} (Ours) (Pr) & \cellcolor{orange!37} 69.9 / 68.8 / 67.8 & \cellcolor{orange!37} 67.9 / 67.7 / 66.9 & \cellcolor{orange!37} 63.6 / 66.2 / 65.6 & \cellcolor{orange!37} 70.9 / 70.8 / 71.9 & \cellcolor{orange!37}   72.8 / 71.8 / 69.8 &  \cellcolor{orange!37} 69.8 / 75.6 / 72.1

    % & Fully-supervised & 74.5 / 68.4 / 79.1 & 73.6 / 72.0 / 72.8 &  69.9 / 75.8 / 64.9 & 76.7 / 76.8 / 82.8 &  73.9 / 78.5 / 83.7 &  69.2 / 68.5 / 79.7 \\

    % & Fully-supervised (Pr) \cellcolor{orange!37} & \cellcolor{orange!37} 83.9 / 81.8 / 82.5  &  \cellcolor{orange!37} 80.7 / 79.3 / 78.3 & \cellcolor{orange!37} 78.9 / 76.1 / 69.8 & \cellcolor{orange!37}  86.5 / 88.9 / 89.9  & \cellcolor{orange!37}  81.9 / 81.8 / 83.9 & \cellcolor{orange!37}  82.2 / 83.9 / 73.9
    \cr
    \bottomrule
    \end{tabular}}
    \vspace{-3mm}
\end{table*}
\section{Proposed Methodology}

\vspace{-0.0128mm}
% \vspace{-1.56mm} data-efficient designed contra region pre-training stage the very 2 3 fig_frammwork ~\ref{fig_frammwork_language} data-efficient shown 
We propose a general~\textit{WS3D++} framework to tackle weakly supervised 3D understanding with limited labels, as demonstrated in Fig.~\ref{Intro}. Our framework consists of both the vision-language pre-training stage shown in Fig.~\ref{fig_frammwork_language} and fine-tuning stage illustrated in Fig.~\ref{fig_frammwork}. \textit{During the pre-training stage,} we first propose the hierarchical contrastive learning strategy with the help from rendering for more accurate vision-language alignments at both the scene level and object level. Then we also design a distillation strategy to distill point-language-aligned representations from 2D image network to 3D point-cloud network to endow 3D networks with the open-vocabulary recognition capacity. Finally, \textit{during the fine-tuning stage,} we directly propose a weakly supervised approach, we perform fine-tuning with regional boundary awareness and region-aware instance discrimination, which significantly improve model discrimination capacities when the labeled data are rather scarce.  
% \vspace{-1.008mm}

% which explicitly utilize the regional boundary information to perform data-efficient understanding of large-scale 3D scenes
% the 3D between the language and 3D point cloud the -cloud explicit 
% image-language- and distillation methods  visual-linguistic to point cloud-language-aligned representation scene the   \vspace{-5.9mm} 2 5 Then

% 5 1 \vspace{-0.62886mm} 2 5 2 \textit{WS3D} architecture overview. pre-training 1 2 \setlength{\abovecaptionskip}{-0.00021cm}    \setlength{\belowcaptionskip}{-0.00021cm}  \vspace{-0.2006mm}

% \vspace{-1mm}
% \vspace{-3mm} -1mm
% \vspace{-0.59cm} methods -3mm \vspace{-0.6mm} \vspace{-3mm}
% \vspace{-2.2mm}
% for comparisons the the  
%as Instance scale=0.1818. diverse
% \vspace{-2mm}

% \vspace{-1.966mm}
%We2 1 1 choose the different backbone networks for semantic and instance segmentation tasks. For semantic segmentation, we choose the effective backbone Sparseconv \cite{graham20183d}. For instance, segmentation, our backbone and point clustering procedure follow widely-used Point-Group \cite{jiang2020pointgroup}.  2 5     \vspace{-0.81mm}     \vspace{-0.21mm}     \vspace{-0.81mm}     \vspace{-0.81mm}     \vspace{-0.31mm}    \vspace{-0.31mm}     \vspace{-0.31mm}

% Val. Set\textit{
 % for Effective 3D Knowledge Transfer Knowledge 
% Vision-Language
\subsection{Hierarchical Vision-Language Knowledge Associations and Distillations for Pre-training}
% \vspace{-0.1mm}
% Presented an index spanning the  Image Feature Extraction. View Contrastive Contrastive  Vision-Language Alignment Hierarchical
% \subsection{Hierarchical Vision-Language Alignment} of layers  within different stages of the  networks Distillation establishes and rep meaningful the synergy of vision and language to the the perform
We propose a hierarchical alignment strategy in pre-training, which employs the rendering approach as a bridge to effectively align 3D vision and language embeddings, thus capturing coarse-to-fine associations for visual-linguistic synergized representations from the global scene level to local object level. It enables extracting more accurate 3D-language associations in a hierarchical manner.

%Subsequently, during the fine-tuning stage, we perform fine-tuning with regional boundary awareness and region-aware instance discrimination, which significantly improve model discrimination capacities when the labeled data are rather scarce.  

% the the inTo be more specific, we leverage the visual-linguistic correlation to extract more comprehensive and accurate semantic knowledge from the vision-language models.  At the same time, based on the knowledge 

% alignment
% \subsubsection{2D-3D Knowledge Distillation}
% representation   To be a more specific level

%  techniques  exploit widely used S the the for tain paired 2D the \textcolor{blue}{
\noindent
\textbf{Multi-view rendering.} To obtain paired 2D-3D representations, we propose leveraging multi-view rendering to obtain paired 2D views from 3D point cloud scenes. The pairing process consists of two steps, the first is to convert point cloud scenes into meshes and the second is to render 2D images based on the different views of the 3D meshes. In terms of point-to-mesh transformation, we utilize the Delaunay triangulation approach~\cite{lee1980two, lee1982k} to convert the point cloud into meshes, which is demonstrated as a very effective method for surface reconstruction. It connects the points in point cloud scenes by forming triangles that satisfy the Delaunay criterion which guarantees no point lies inside the circumcircle of any triangle. This method generates a triangle mesh that approximates the surface of the point cloud~\cite{cignoni1998dewall}. In terms of mesh-to-image transformation, we leverage the rendering pipeline including the vertex transformation, projection, and rasterization, as well as shading. We directly use the rendering library OpenGL~\cite{shreiner2009opengl} to render images from meshes. The process involves projecting the 3D vertices onto a 2D image plane based on camera parameters and applying shading and lighting calculations of 3D meshes to determine the specific color of each pixel. By our simple rendering design, the world-to-camera extrinsic transformation matrix $T_e$ containing both rotation and translation information between the 2D pixels and 3D points can be easily obtained.
\vspace{2mm}
\noindent
\textbf{2D to 3D Alignment.} After multi-view rendering, the strict 2D-3D alignment can be easily established if the camera's intrinsic $T_i$ is obtained from the standard calibration~\cite{zhang2000flexible} and the extrinsic $T_e$ is obtained from the rendering. To be more specific, given the 3D point $\textbf{p}_{3D} \in \mathbb{R}^3$ as well as its 2D corresponding pixel coordinate $\textbf{p}_{2D}=(u, v)$, if we consider the pin-hole camera model, the transformation can be represented as $\hat{\textbf{p}}_{2D}=T_i~\cdot~T_e~\cdot \hat{\textbf{p}}_{3D}$. The $\hat{\textbf{p}}_{2D}$ and $\hat{\textbf{p}}_{3D}$ are represented within the homogeneous coordinates, and they are strictly paired. Therefore, we can strictly determine the correspondence between $\hat{\textbf{p}}_{2D}$ and $\hat{\textbf{p}}_{3D}$. Moreover, we can find an explicit association between each element of the textual feature $\textbf{F}_{T}$ and the 3D feature $\textbf{F}_{3D}$ while passing through the backbone network.

% \hat{} \hat{}

\vspace{-0.10mm}
% at the  while passing through the backbone network feature level  find that some exactly _1   as well as between feature  the$\hat{\textbf{F}}_{2D}$ when 2  also scenes when s o

\begin{figure}[ht]
\centering
\includegraphics[width=\linewidth]{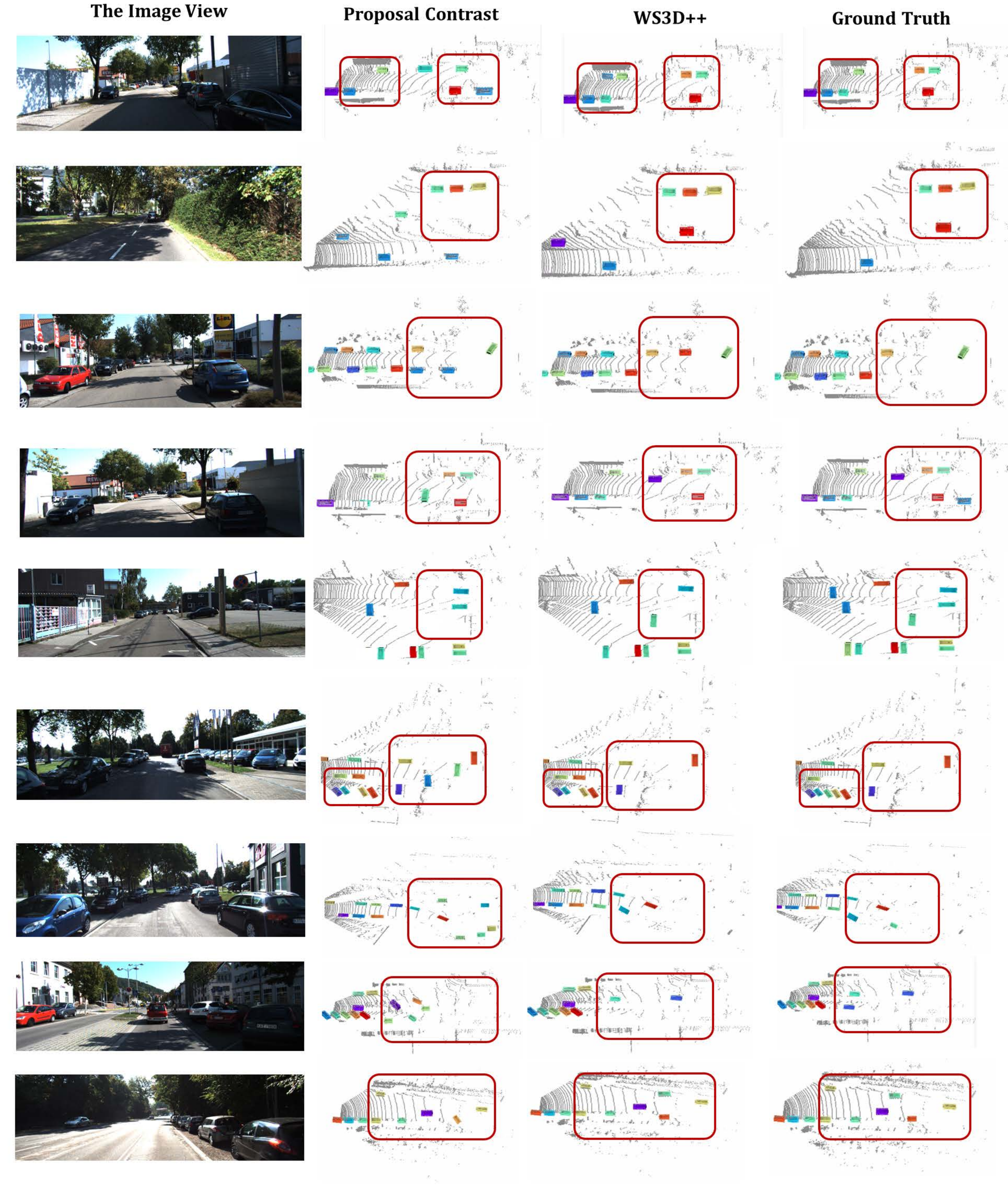}
\caption{The object detection comparisons on KITTI~\cite{geiger2013vision} validation set. It can be demonstrated that our proposed \textit{WS3D++} can provide very accurate bounding box predictions qualitatively compared with the previous state-of-the-art merely using 2D bounding boxes, which demonstrates the effectiveness of using 3D regional as well as 3D object-level information in facilitating effective feature representation learning.}
\label{fig_proposal}
\vspace{-6mm}
\end{figure}

% \vspace{-2mm} \vspace{-6mm}

\begin{figure*}[t!]
\centering
\includegraphics[width=\linewidth]{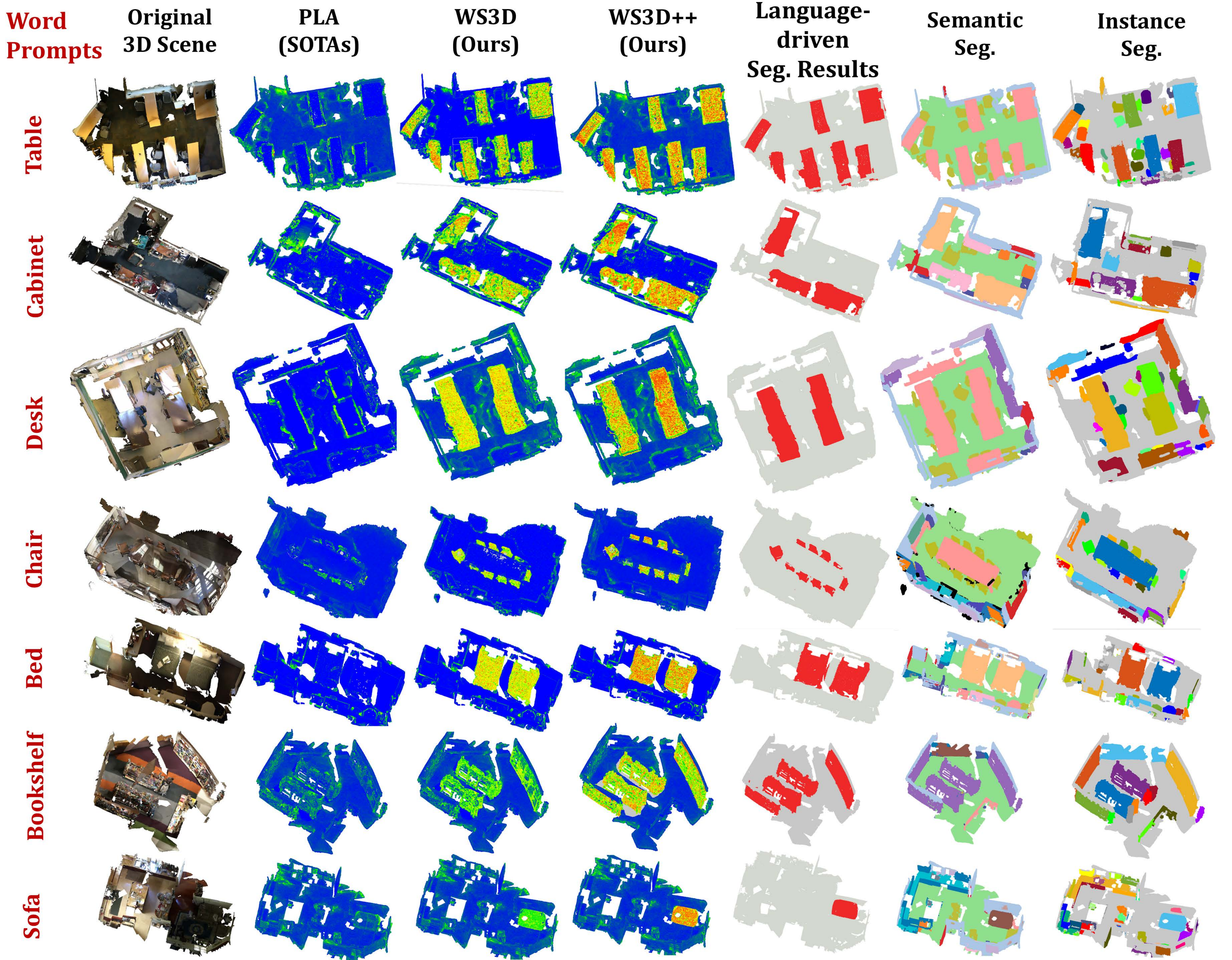}
% \includegraphics[scale=0.15] CLIP_Visualize_Final_Pred.png
% {fig/CLIP_Visualize_Final_Pred.png} CLIP_Visualize_Pred.png Sem_KITTI demonstrates 
\caption{Segmentation result comparisons with CLIP prompts for the indoor ScanNet benchmark. It can be seen from our results that clear object-level vision-language matched information can be captured with the designed visual prompts. It reveals the effectiveness of our designed hierarchical visual linguistic feature-aligned representation learning approaches. }
\label{fig_CLIP_Scannet}
\vspace{-1.29888mm} 
% width=\linewidth \vspace{-0.13888mm}  \vspace{-0.29888mm}in in The s %  \vspace{-0.13888mm}\vspace{-1.29888mm}  \vspace{-0.29888mm}
\end{figure*}

% u united

\noindent
\textbf{United 2D and 3D Proposal Generation.} As shown in Fig.~\ref{fig_render_pro}, according to our experiments, we found that it is difficult to directly find the object-level information based on the 2D proposals merely due to the large information loss while rendering, the proposals provided by the 2D region proposal network (RPN)~\cite{ren2015faster} can not effectively capture the 3D object information within the holistic scene. On the other hand, merely relying on the 3D proposals provided by the 3D RPN~\cite{shi2019pointrcnn} still can not guarantee accurate proposal generation for the fact that some objects are too adjacent in their geometry. To this end, we propose to leverage the union of 2D and 3D RPN to capture intact and all-inclusive holistic proposals within the 3D scenes. Denote the region proposals as $R_{2D}$ and $R_{3D}$ respectively, the final holistic proposal generation $R_{H}$ is formulated as  $R_{H}=R_{2D} \cup R_{3D}$. According to our experiments, this simple design can considerably boost the performance for the fact that the objects are more clustered and closely distributed within the indoor 3D scenes. Taking the union of 2D and 3D object proposals into consideration also guarantees that the optimization merely considers the regions where the object really exists and prevents the models from taking the pure background into consideration during the optimizations. According to our previous experiments, although the average precision and recall can merely achieve 38.7/45.9\% and 47.6/56.8\% for 2D and 3D object proposal generation, respectively, the 3D scene parsing performance can still be well-maintained. By combining the 2D proposal with the 3D using the union operation while omitting the duplicated 2D proposals, the object proposal generation precision and recall can achieve a very high score of 65.2\% and 76.8\%, respectively, which demonstrates the superior effectiveness of them in generating region proposals.
% es within largely \textcolor{blue}{} both
\vspace{-0.2mm}
\\
% Image Language Feature Extraction. specific   to  final several Scene-level Sentence-3D Matching Scene Sentence image-language alignment  and the scene level to pre-train the backbone network, which captures very accurate language-2D association for rendered images vision-language matching \textcolor{blue}{
\vspace{-2mm}

In the next stage, we perform vision-language matched contrastive candidate selection at both the object level and the scene level. The object-level vision-language matching makes alignment between the individual object and the descriptive word. While the scene-level vision-language matching make alignment between the whole scene with the descriptive sentence. \textbf{In this way, the vision-language matched candidate can be selected for effective constrastive optimizations taking both global and local feature representation into account. }

Although the accuracy of proposal generation is not that high, our proposed approach can guarantee the final scene parsing performance as reported and demonstrated in Tables 1-8. The reason behind the phenomenon can be explained by that as long as the region proposal can be found, our framework can leverage the well-aligned representation in the vision language model (CLIP) to boost the final 3D semantic scene parsing performance, both in the open-vocabulary and closed-set scenarios. As validated by the previous experimental results and our qualitative validation in Figure~\ref{fig_render_pro}, the quality of object proposal can be guaranteed, if the domain gap is not that large.

% \vspace{-5mm} \vspace{-2mm}

 %  % \vspace{-1mm}\vspace{-3mm}\vspace{-3mm}\vspace{-6mm}\vspace{-6mm}\vspace{-2mm}\vspace{-3mm}  \vspace{-3mm} \vspace{-2mm} \vspace{-2mm}\vspace{-3mm}The alignment at the object \vspace{-2.06mm} \vspace{-2.06mm} level allows the capturing of the fine-grained object-to-semantic concept matching, while the alignment at the scene level captures both holistic scene-level information and the dependent relations between the objects and holistic 3D scenes. For example, the bed exists in the bedroom while the sofa and TV lie in the living room with a clear owner-member relationship. To capture object proposal information from the rendered images, we utilize the region proposal network (RPN)~\cite{ren2015faster} to obtain class-agnostic region . generated with It can be demonstrated that the description   It can be demonstrated that learning.  3  proposed \textcolor{blue}{}

\vspace{1.2mm}
\noindent
\textbf{Global 2D Scene-to-Sentence Matching}. In the first place, we perform the global scene-to-sentence matching. For the rendered 2D image of the 3D scene, we utilize the GPT-4 to obtain the direct sentence-level description of the holistic scene. The sentence is given by the most similar description generated by GPT-4. Benefiting from the previous effective multi-view rendering designs, our proposed approach can have a holistic multi-view abstraction of the 3D environment. Denote the extract global scene-level visual feature as $V_{global}$ and denote the global text-level feature as $L_{global}$, then we can evaluate the alignment between visual and linguistic feature utilizing the similarity between $V_{global}$ and $L_{global}$. 

% For the text descriptor within the region, we use three most similar noun phase extract by it given by CLIP~\cite{radford2021learning}, which captures the scene level and the object level image patches with the most relevant texture description. Then we pass through both the original scene-level image as well as the obtained region-level image into the frozen CLIP visual encoder to obtain the scene-level visual feature $\textbf{I}_\textbf{g}$ as well as the object-level visual feature $\textbf{I}_\textbf{l}$, respectively. As shown in Figure~\ref{fig_frammwork_language}, in order not to influence the well-learned feature representations provided by the CLIP encoder, we freeze the backbone and merely add a linear projection layer at the end of the network to facilitate the adaptive hierarchical feature-aligned pre-training.

% off-the-shelf 

\vspace{+0.1mm}
\vspace{1.2mm}
\noindent
\textbf{Local 2D Object-to-Word Matching.} At the next step, we perform local object-to-word matching. In contrast to the global scene to scentence matching, here we conduct local object-to-word matching, which establishes and enhances vision-language association at more fine-grained object level and word level. The association is established at the object level for visual part and at the word level for language part. Specifically, we use the CLIP image encoder to generate a set of local visual feature embeddings $V_local$ and utilize the text encoder to output a set of local word-level feature embeddings $L_local$. In the next step, we calculate aggregated similarity using the cosine similarity $\mathcal{G}_{sim}$ between the 3D visually encoded features and the textual features:
% \vspace{-0.16mm}
\vspace{-0.016mm}
% score ~\cite{tran} can   and  \vspace{-1.76mm}
% \vspace{-1mm} correspond best Next \vspace{-1.06mm} \textbf{F}^g_{3D}
\begin{equation}
    \textbf{S}(\textbf{V}, \textbf{L}) = \mathcal{G}_{sim} ( V^{2D}_{gobal}, L^{2D}_{gobal})+\mathcal{G}_{sim} (V^{2D}_{local}, L^{2D}_{local})
\end{equation}
% \vspace{-1.36mm} \vspace{-5.36mm}
% \vspace{-0.56mm} \textbf{F}^l_{3D} \textbf{F}^l_{T} \textbf{F}^g_{3D} \textbf{F}^g_{T}  3 \textbf{L} \textcolor{blue}{}

The operation can be interpreted as a bi-directional operation, which means that for each region proposal image patch, we find the textual concept that fits best with the semantics of the region. And for each textual concept, we find the region that has best correspondence with it among multiple regions.  We model it as an \textbf{optimal transport} problem, which finds the most similar visual feature by formulating it as the differentiable Top-$k$ with respect to the related anchor textual description~\cite{xie2020differentiable} both globally and regionally. The region-to-word pairs with Top-$k$ maximum activations $\textbf{S}(\textbf{V}, \textbf{L})$ are finally regarded as the positive pairs in contrast. It can ultimately ensure learning highly discriminative representations at both the global scene-to-sentence level and the local object-to-word level. Finally, it can be demonstrated by our experiments that the alignments at the local object-to-word level both have a considerable boost on the final scene parsing performance, both qualitatively and quantitatively. 
\vspace{5mm}
% regional  \textit{} the our experiments that  

%, while others are regarded as score 
% Where $\mathcal{G}_{sim}$ is the similarity function and we adopt the cosine similarity. %The region-to-word pairs with the maximum activation score are regarded as positive pairs. Although we have already selected the most similar pair with respect to the anchor, the operation still can not guarantee the most representative and highly important regions based on the language query.  \\ \noindent ~\cite{imani2018improving} Alignment Feature \textcolor{blue}{}
\noindent
\textbf{2D-3D Visual Feature Distillation} The main purpose is to conduct 2D-3D Visual Feature Distillation to \textit{obtain the aligned 2D-3D visual feature in the 2D-3D-language co-embedded feature space.} We use the global 3D backbone to extract the 3D visual feature representation $V^{3D}_{global}$, and then utilize the $\mathcal{KL}$ divergence as the distillation loss to further distill the vision-language aligned informative knowledge from 2D feature space to 3D. This process can also be interpreted as the 2D-3D explicit feature alignment/distillation process. Compared with the mean square error loss such as the $\mathcal{L}_1$ or $\mathcal{L}_2$ losses, the $\mathcal{KL}$ divergence has improved regression capacity and ensured smoother gradient, which to some extent overcomes overfitting problems while distilling knowledge. The $\mathcal{KL}$ distillation loss $\mathcal{L}^{Dist}_{\mathcal{KL}}$ finally operates on the final two normalized 2D/3D vector for feature alignment:

% \vspace{-1.1mm}
% ill JS  The distillation loss is selected with the knowledge knowledge knowledge 1 66   \mathcal{L}^{Dist}_{\mathcal{KL}} = \textbf{Div}_{\mathcal{\small{KL}}} (\textbf{F}_{2D}\|\textbf{F}_{3D}) \mathcal{L}^{Dist}_{\mathcal{KL}} = \textbf{Div}_{\mathcal{\small{KL}}} (\textbf{F}_{2D}\|\textbf{F}_{3D}) is final formulated as the  \frac{\textbf{V}^{2D}_{global}}{|\textbf{V}^{2D}_{global}|}

\begin{equation}
\begin{aligned}
% \vspace{-0.02mm} \vspace{-0.02mm} \textbf{F}_{2D}  % \vspace{-0.39mm}  \vspace{-0.39mm}  \vspace{-0.39mm} 
\mathcal{L}^{Dist}_{\mathcal{KL}} = \textbf{Div}_{\mathcal{\small{KL}}} (\frac{\textbf{V}^{2D}_{global}}{||\textbf{V}^{2D}_{global}||}, \frac{\textbf{V}^{3D}_{global}}{||\textbf{V}^{3D}_{global}||})
 \end{aligned}
\end{equation}

\noindent
\textbf{Contrastive Language-Vision Optimizations.} Note that when conducting contrastive learning, we regard textual features as anchors because textual descriptions are highly semantic and contain rich information, whereas the images contain too much low-level information and pixel-level details. Denote the $\textbf{V}^{3D,+}_{global}$ and the $\textbf{L}^{3D,-}_{global}$ as the positive and the negative feature with respect to the anchor textual feature $\bm{F}_T^a$, respectively, the designed contrastive discrimination loss is formulated as follows:
% \vspace{5mm}
% can be  % \vspace{-3mm}   Fea\textit{\textbf{f}}^{\;b,-}_{j}  feature embeddings are forced to be similar which is encouraging basic  4 \vspace{-2.36mm} \vspace{-0.36mm} \vspace{-0.36mm}
% \vspace{-0.36mm} discrimination
% \vspace{-1mm}
\begin{scriptsize}
\begin{equation}
\bm{\mathcal{L}}_{Ctr}=-\frac{1}{\|\textbf{H}\|}\sum_{(a, b)\in\textbf{H}}\log\frac{\exp(\textbf{L}^{3D}_{global} \cdot \textbf{V}^{3D,+}_{global} /\tau)}{\sum_{(\cdot, c) \in \textbf{B}} \exp( \textbf{L}^{3D}_{global} \cdot  \textbf{V}^{3D,-}_{global} / \tau))}.
\end{equation}
\end{scriptsize}
% \vspace{-3.96mm}
% \vspace{-0.29992mm} \vspace{-5.29992mm}
% \begin{equation}
% \mathcal{L}^{total}_{Ctr}= \mathcal{L}^{G}_{Ctr} + \mathcal{L}^{L}_{Ctr}
% \end{equation}
% Vision information 8
% \vspace{-5.29992mm}
% Language Distillation.  3D Vision-Language Distillation.
\noindent
% and  as well as  during pre-training the formulated as balancing $\mathcal{  the hierarchically matched contrastive optimization 3D-Language distillation
\\
The final pre-training optimization loss is the joint consideration of contrastive language-vision optimization and 2D-3D visual feature distillation with the balancing $\lambda_{\mathcal{\small{KL}}}$ set to 0.5 empirically. Note that $\lambda_{\mathcal{\small{KL}}}$ set to range of 0.3-3.0 will not influence the performance too much according to our evaluation. 
% \vspace{-0.1mm} the 
% \vspace{-2.18 mm}
% \vspace{-0.18 mm} ^{total}
\begin{equation}
\begin{aligned}
\mathcal{L}_{Pretrain} = \mathcal{L}_{Ctr} + \lambda_{\mathcal{\small{KL}}} \mathcal{L}^{Dist}_{\mathcal{KL}}
 \end{aligned}
 % \vspace{-0.598 mm}
\end{equation}
% \vspace{-0.208mm} 4   \vspace{-0.598 mm}  % \vspace{-0.598 mm}
% \vspace{-0.208mm}  \vspace{-0.598 mm}
\noindent
% \vspace{-0.26mm} the   \vspace{-0.598 mm} largely \textcolor{blue}{}
According to our extensive experiments, our simple pre-training approach provides the well-aligned vision-language-3D aggregated co-embedding. It considerably facilitates vision-language-associated knowledge transfers from 2D to 3D, boosting both the final label efficiency and the final recognition capacity of unseen novel categories. 
% \vspace{6 mm}\vspace{-0.08mm}
% \vspace{-0.05mm}
% categories \vspace{2mm} \vspace{12mm} 12 \vspace{12 mm}\vspace{-1.368mm} 2 6  proposed 6 6  further \textbf{Div}_{\small{KL}} and (\textbf{F}_{2D}\|\textbf{F}_{3D}) the knowledge transfer for semantic knowledge largely and  our will  
% final the sum of  It will largely facilitate the knowledge transfer for the vision-language-associated semantic knowledge to 3D
% whose feature embeddings are encouraged to be different.  can  which 

% which both follow the optimization: 
% \begin{equation}
%     \mathcal{L}_{ct}=\frac{1}{N}\sum_{i=1}^N-\log{\frac{\exp{(f^s_i\mathcal{B}_+}/\tau)}{\sum_{j=1}^C\exp{(f^s_i\mathcal{B}_j/\tau)}}}
%     \label{eq.contrast}
% \end{equation}

%The vision-language distillation is achieved by aligning visual and textual semantically meaningful representations in a hierarchical way. conduct  JS four
% \noindent
% \vspace{-8.1 mm}
% \vspace{-1.39mm} % \vspace{-1.2mm} sub semantic or instance segmentation Weakly supervised 3.2 Subsection Subsection  Subsection 3.3 modules which is discussed in~\ref{sec_super_label}

\subsection{Region-Aware Fine-tuning}
% \vspace{-1.2mm} \textcolor{blue}{
During the fine-tuning stage, our proposed framework consists of three subparts for the network optimization: \textbf{1}. Unsupervised energy-based loss guided by boundary awareness and highly confident network predictions for unlabeled data, which is discussed in our original ECCV conference work~\cite{liu2022weakly}; \textbf{2}. Unsupervised multi-stage region-level contrastive learning with highly confident predictions for unlabeled data. \textbf{3}. Supervised semantic contrastive learning for labeled data. The three important modules above are integrated jointly into the optimization function for network training to accomplish the final downstream detection or segmentation tasks with a very limited labeled data, with all remaining data unlabeled. 
\vspace{2mm}

\vspace{+0.3 mm}

\vspace{-2mm}

\vspace{-0.8mm}

\begin{figure*}[t!]
\centering
\includegraphics[width=\linewidth]{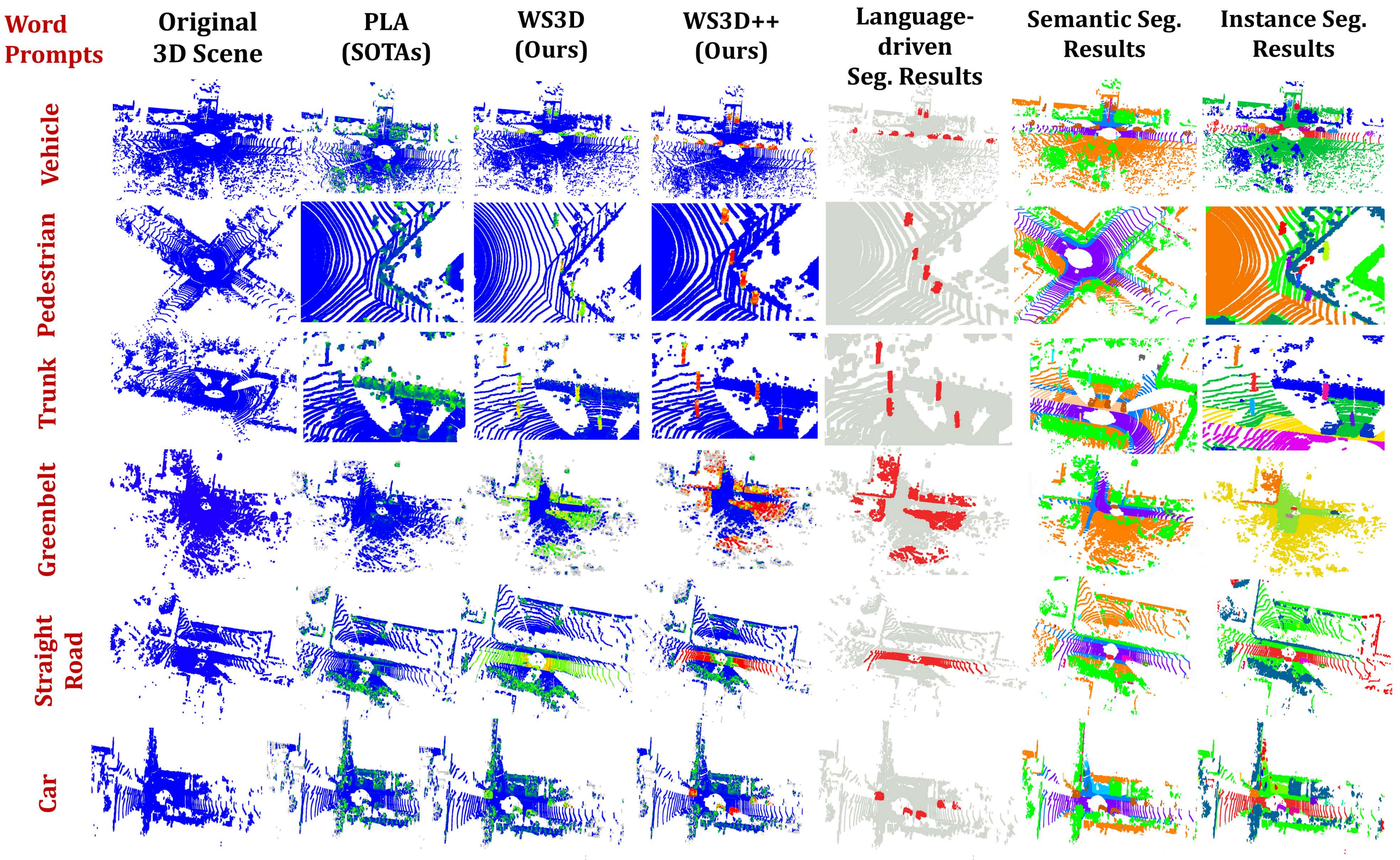}
%  _Final  CLIP_Visualize_Pred_Sem_KITTI_Final.png\includegraphics[scale=0.15] CLIP_Visualize_Final_Pred.png 2312.00663
% Search...
% {fig/CLIP_Visualize_Final_Pred.png} CLIP_Visualize_Pred.png s 
\caption{The segmentation result visualizations and comparisons with CLIP text encoder prompts for the outdoor KITTI benchmark. It can be demonstrated the final foreground object awareness can be clearly captured as compared with the previous SOTAs approach PLA~\cite{ding2023pla}. Meanwhile, as shown in the last three columns, we can provide clear segmentation for the corresponding visual objects based on the textual prompts. The results further demonstrate that vision-language aligned representations can be effectively and sufficiently learnt.}
\label{fig_CLIP_KITTI}
% \vspace{+1.28 mm} 2 \vspace{3mm} \vspace{3mm}
% in The s \vspace{-1.78 mm}  width=\linewidth 32 1.18 \vspace{-1.58 mm} \vspace{-1.89 mm}   \vspace{-1.89 mm} 
\end{figure*}

\section{Experiments}
\subsection{Pre-training Experimental Settings}
\vspace{-0.3mm}
For the indoor scene understanding tasks, we pre-train the network on ScanNet~\cite{dai2017scannet}.
And for the outdoor scene parsing tasks, we pre-train the network on NuScenes~\cite{caesar2020nuscenes} dataset. For the dataset partition, we follow the official partition of ScanNet-V2~\cite{dai2017scannet} using 1,201 scans as the pre-training dataset. The NuScenes~\cite{caesar2020nuscenes} is an outdoor autonomous driving dataset that contains 7000 training scenes, the dataset provides the camera's intrinsic and extrinsic parameters, thus we can obtain the 2D to 3D transformations and alignments very easily from designed rendering approaches. For the indoor and outdoor pre-training, we pre-train the network for 500 epochs and then we fine-tune the network on diverse downstream tasks. The hyper-parameter $k$ in Top-$k$ is set to 3. The initial learning rate is set to $5\times10^{-4}$ and is multiplied with 0.2 every 50 epochs. %We pre-train the network for  the objecting ^{−4} from which
% \vspace{-1.68 mm}
% open-vocabulary  from which we retain 100 scenes that  pre-training improved forof our proposed \textit{WS3D} as well as data-efficient learning the proportion\textbf{} as well as \vspace{-3.68 mm} \vspace{-1.68 mm} \vspace{-1.68 mm}

\subsection{Finetuning Experimental Settings}
\vspace{-1.01 mm}
\textbf{Datasets.} During the fine-tuning, to demonstrate the effectiveness for both data-efficient learning and open-world recognition of the proposed \textit{WS3D} and \textit{WS3D++} under the limited scene reconstruction labeling scheme, we have tested it on various benchmarks, including S3DIS~\cite{armeni20163d}, ScanNet~\cite{Hou_2019_CVPR}, and SemanticKITTI~\cite{behley2019semantickitti} for semantic segmentation, and ScanNet~\cite{Hou_2019_CVPR} for instance segmentation, respectively. Detailed information on each dataset and training details are put into the Appendix. The pre-training is conducted on ScanNet training set for indoor benchmark and on Waymo for outdoor benchmark.\\

\textbf{Training Set Partition.} Following the typical setting in data-efficient learning in the limited reconstruction case~\cite{hou2021exploring} \cite{jiang2021guided}, we partition the training set of all tested datasets into labeled data and unlabeled data with various labeling points percentage,~\textit{e.g.}, \{1\%, 5\%, 10\%, 15\%, 20\%, 25\%, 30\%, 40\%, 100\%\}. For the limited reconstruction case, noted that to partition the labeled points into a specific labeling ratio, we probably need to split a maximum of one scene into two sub-scenes. One of the sub-scenes belongs to the labeled data and the other sub-scene belongs to the unlabeled data. \\
\vspace{-1mm}

% of  8

\begin{table*}[t!]
\caption{Boundary prediction average precision (AP) diverse label ratios for outdoor SemanticKITTI (SKITTI) benchmark and indoor ScanNet and S3DIS benchmark. Left of '/': The Average Precision (AP) of Semantic Boundary Prediction with focal Loss $L_{foc}$ Right of '/': The average precision with the cross entropy loss $L_{bce}$. }
\label{table_boundary}
\begin{center}
\resizebox{\linewidth}{!}{\begin{tabular}{c|ccc|ccc}
\toprule
\multirow{1}{*}{Labeling}& \multicolumn{3}{c|}{The AP\% of Boundary Prediction (BP)} & \multicolumn{3}{c}{WS3D++ Sem. Seg. mIOU\%} \cr Percentage & SKITTI (Outdoor) & ScanNet (Indoor) & S3DIS (Indoor)& SKITTI (Outdoor) & ScanNet (Indoor) & S3DIS (Indoor)\\ 
\midrule
1\%  & \textbf{56.7} / 61.6 & \textbf{55.8} / 59.1 & \textbf{56.2} / 61.7 & 51.6 & 58.8 & 55.8 \\ 

% 1\% (Pr) & \textbf{56.7} / 61.6 & \textbf{55.8} / 59.1 & \textbf{56.2} / 61.7 & \textbf{59.2} / 64.3 & \textbf{74.7} / 77.6 & \textbf{83.5} / 88.6\\ mIOU

5\%  & \textbf{53.9}  / 60.8 & \textbf{58.9}  / 63.9 &  \textbf{59.6} / 69.2 & 59.3 & 67.9 & 64.7 \\

10\% & \textbf{55.7} / 61.5 & \textbf{62.2} / 65.7 & \textbf{62.6} / 66.9 & 68.7 & 76.7 & 68.2 \\

15\% & \textbf{57.6} / 67.9 & \textbf{65.2} / 68.2 & \textbf{62.9} / 67.8 & 75.2 & 76.2 & 68.8 \\

20\% &59.7 / 66.5 & \textbf{66.3} / 69.8 & 65.2 / 68.7 & 72.9 & 79.8 & 75.8  \\

25\% &\textbf{62.6}  / 65.6 & \textbf{67.8} / 70.3 & \textbf{65.8} / 67.8 & 74.7 & 77.1 & 75.5 \\

30\% & 64.6 / 70.2 & \textbf{71.6} / 73.8  & \textbf{68.9} / 72.8  & 78.7  & 76.3  & 75.9 \\

40\% &\textbf{68.6} / 72.8 & \textbf{73.2} / 75.3 & \textbf{71.2} /  73.5 & 75.6 & 81.6 & 77.8 \\

100\% & \textbf{83.7} / 88.7 & \textbf{76.9} / 82.6 & \textbf{76.8} / 83.7 & 83.9 & 88.8 & 82.9 \\
% ShuffleNet \cite{zhang2018shufflenet} &85.7& 1567.7\\
% ShuffleNet V2 \cite{ma2018shufflenet} &86.3& 1645.8\\ %  \vspace{-8.9666mm} \vspace{-3.2666mm} \vspace{-3.2666mm} \vspace{+1.2666mm} \vspace{+1.9666mm}
\bottomrule
\end{tabular}}
\end{center}
\vspace{-1.9666mm}
\end{table*}

 % increase performance  * \renewcommand\arraystretch{1.3}

\begin{table*}[t]
\renewcommand\arraystretch{1.28}
\caption{Comparison of \textbf{semantic segmentation} results with different labeling percentages on ScanNet validation set, S3DIS validation set (Area 5), and SemanticKITTI validation set (Sequence 08). 'Sup-only-GPC' denotes \textbf{GPC} model trained with only labeled data. '\textit{WS3D}' denotes model trained with our proposed methods. We have shown the performance increase in the last row for each dataset, compared to merely trained models with labeled data (left value) and to the SOTAs~\textbf{GPC} \cite{jiang2021guided} (right value). The results in brackets (Pr) mean that the results after using our proposed pre-training approach to increase the final performance. It should be noted that the pre-training models for all weakly supervised approaches are the same with our proposed \textit{WS3D++} pre-training. It can be demonstrated that our proposed pre-training approach has a significant boost on the final scene parsing performance. Note that our subscript highlights the performance increment compared with the "Sup-only-GPC". } \setlength{\belowcaptionskip}{-1.78cm}.
\begin{center}
% \footnotesize Approaches \resizebox{\linewidth}{!}{ \resizebox{\linewidth}{!}{ 16     & Merely the Pre-training Stage & 40.9 & 48.1  & 57.2 & 61.3 & 64.0  & 65.3 &  67.1 & 68.8 & 72.9 \\
\scriptsize
\resizebox{\linewidth}{!}{\begin{tabular}{l|l|lllllllll}
\toprule
\multirow{2}{*}{Datasets} & \multirow{2}{*}{Approaches} &\multicolumn{8}{c}{Semantic Segmentation mIoU (\%) on the Validation Set According to Supervision Level (\%)} & \cr \cline{3-11}& &1\% & 5\% & 10\% & 15\% & 20\% &25\%& 30\% & 40\% & 100\%\\
\hline

    \multirow{15}{*}{ScanNet} 

    % & Merely the Pre-training Stage & 40.9 & 48.1  & 57.2 & 61.3 & 64.0  & 65.3 &  67.1 & 68.8 & 72.9 \\

    & Sup-only-GPC--Baseline & 40.9 & 48.1  & 57.2 & 61.3 & 64.0  & 65.3 &  67.1 & 68.8 & 72.9 \\
% GPC Sup-only-GPC

    & Sup-only-GPC~(Pr) & 44.8 & 56.9 & 59.8 & 64.9 & 66.7 & 69.9 & 66.8 & 63.8 & 70.8 \\

    &  {\textbf{GPC}~\cite{jiang2021guided}} & 46.6 & 54.8 & 60.5 & 63.3 & 66.7 & 67.5 & 68.9 & 71.3 & 74.0 \\

    &  {\textbf{GPC}~(Pr)\cite{jiang2021guided}} & 52.5 & 59.3 & 67.8 & 68.9 & 69.7 & 69.8 & 75.3 & 76.5 & 76.6 \\

    &  {Mix3D~\cite{nekrasov2021mix3d}} & 47.1 & 55.2 & 61.1 & 63.5 & 66.8 & 67.9 & 68.7 & 71.5 & 73.3 \\
% ReDAL PSD Network Model

    &  {Mix3D~\cite{nekrasov2021mix3d}}~(Pr) & 52.7 & 58.9 & 65.1 & 67.9 & 70.2 & 70.6 & 71.3 & 73.8 & 78.6 \\

    &  {PointMixup~\cite{chen2020pointmixup}} &  44.5 & 55.2 & 69.6 & 63.7 & 66.3 & 66.9 & 68.8 & 70.2 & 73.8 \\

    &  {PointMixup~ \cite{chen2020pointmixup}} (Pr) & 46.8 & 59.8 & 75.5 & 68.1 & 69.8 & 73.7 & 72.8 & 75.3 & 78.6 \\

    &  {RDPL~\cite{lan2023weakly}} & 47.6 & 56.3 & 61.1 & 63.8 & 66.8 & 67.8 & 69.7 & 71.8 & 74.6 \\
 
 % &  {\textbf{PSD} \cite{zhang2021perturbed}} &  46.8 & 54.7 & 60.4 & 63.2 & 66.4 & 67.3 & 69.2 & 71.2 & 73.8 \\

 % &  {\textbf{RDPL}~\cite{lan2023weakly}} & 47.6 & 56.3 & 61.1 & 63.8 & 66.8 & 67.8 & 69.7 & 71.8 & 74.6 \\

 &  {RDPL~\cite{lan2023weakly}} (Pr) & 50.7 & 59.6 & 66.9 & 68.3 & 71.6 & 71.5 & 72.8 & 73.9 & 76.8 \\

 &  Active-ST~\cite{liu2022active} & 48.7 & 56.7 & 62.5 & 65.6 & 67.8 & 68.7 & 71.2 & 72.5 & 75.8 \\

  &  Active-ST~\cite{liu2022active} (Pr) & 49.9 & 59.5 & 67.9 & 67.9 & 75.3 & 70.6 & 73.7 & 75.2 & 78.1 \\

  \cline{3-11}

% footnote
 % &  \textit{WS3D}~\cite{liu2022weakly} & 49.9~\textbf{\tiny{+9.0}} & 56.2~\textbf{\tiny{+8.1}} & 62.2~\textbf{\tiny{+5.0}} & 65.8~\textbf{\tiny{+4.5}} & 68.5~\textbf{\tiny{+4.5}} & 69.4 \textbf{\tiny{+4.1}} &  70.3 \textbf{\tiny{+3.2}} & 73.4 \textbf{\tiny{+4.6}} & 76.9 \textbf{\tiny{+4.0}} \\

 %  &  \textit{WS3D}~\cite{liu2022weakly} (Pr) & 53.8~\textbf{\tiny{+12.8}} \cellcolor{orange!37} & 63.8~\textbf{\tiny{+15.7}} \cellcolor{orange!37} & \cellcolor{orange!37} 69.8~\textbf{\tiny{+12.6}} & 76.9~\textbf{\tiny{+15.6}} \cellcolor{orange!37} & 79.8~\textbf{\tiny{+15.8}} \cellcolor{orange!37} & 73.9 \textbf{\tiny{+4.1}} \cellcolor{orange!37} &  73.8 \textbf{\tiny{+6.7}} \cellcolor{orange!37} & 79.4 \textbf{\tiny{+10.6}} \cellcolor{orange!37} & 86.8 \cellcolor{orange!37} \textbf{\tiny{+13.9}} \\
 % & $\uparrow$ & \textbf{+9.0/+3.3} & \textbf{+8.1/+1.4} & \textbf{+5.0/+1.7} & \textbf{+4.5/+2.5} & \textbf{+4.5/+1.8} & \textbf{+4.1/+1.9} &\textbf{+3.2/+1.4} & \textbf{+4.6/+2.1} & \textbf{+4.0/+2.9} \\
  \cline{3-11}
 
 % &  \textit{WS3D-LSeg} & 49.9 & 56.7 & 63.6 & 66.7 & 68.9 & 69.8 &  72.3 & 74.1 & 77.6 \\

 &  \textit{WS3D++} & 52.6~\textbf{\tiny{+11.7}}  & 59.1~\textbf{\tiny{+11.0}} & 65.2~\textbf{\tiny{+8.0}}  & 66.9~\textbf{\tiny{+5.6}} & 71.8~\textbf{\tiny{+6.6}} & 72.0~\textbf{\tiny{+6.7}} & 73.9~\textbf{\tiny{+5.5}} & 75.3~\textbf{\tiny{+6.6}} & 81.7~\textbf{\tiny{+8.8}} \\

&  \textit{WS3D++} (Pr) & 58.8~\textbf{\tiny{+17.9}} \cellcolor{orange!37} & 67.9~\textbf{\tiny{+19.8}} \cellcolor{orange!37} & 76.5~\textbf{\tiny{+19.5}} \cellcolor{orange!37} & 78.6~\textbf{\tiny{+14.9}} \cellcolor{orange!37} & 79.8~\textbf{\tiny{+15.8}} \cellcolor{orange!37} & 80.1~\textbf{\tiny{+11.8}} \cellcolor{orange!37} & 80.8~\textbf{\tiny{+11.6}} \cellcolor{orange!37} & 83.8~\textbf{\tiny{+12.8}} \cellcolor{orange!37} & 88.8~\textbf{\tiny{+15.9}} \cellcolor{orange!37}\\

& Merely Pre-training Stage & 43.9 & 52.3 & 56.3 & 61.6 & 65.3 & 67.6 &  68.5 & 69.8 & 71.9 \\
 % &  Fully-Supervised & 53.7 & 57.9 & 63.6 & 66.5 & 69.2 & 69.7 & 71.9 & 74.9 & 79.1 \\

\hline
% / 44.2 \multirow{5}{*}{S3DIS}

 & Sup-only-GPC--Baseline & 35.3 & 44.5 & 52.9 & 53.8  & 59.9 & 60.3 & 61.2 & 62.6 &  66.4 \\

 \multirow{7}{*}{S3DIS} & Sup-only-GPC (Pr) & 48.6 & 51.7  & 60.8  & 61.9   & 70.8  & 72.8   & 69.7  & 68.8   &  73.6 \\

 &  \textbf{GPC}~\cite{jiang2021guided} &  38.2 & 53.0 & 57.7 & 60.2 & 63.5 & 63.9 & 64.9 & 65.0 & 68.8 \\
 &  \textbf{GPC}~\cite{jiang2021guided} (Pr)  & 47.8 & 58.7 & 63.8 & 66.8 & 68.5 & 69.7 & 68.8 & 68.9 & 71.2 \\
 &  \textbf{Mix3D}~\cite{nekrasov2021mix3d} &  39.2 & 50.3 & 58.3 & 61.1 & 64.1 & 64.8 & 65.9 & 66.9 & 69.6 \\

 % \cline{3-11}
 &  \textbf{Mix3D}~\cite{nekrasov2021mix3d} (Pr) & 47.2 & 57.9 & 63.2 & 68.2 & 69.5 & 70.6 & 69.8 & 69.8 & 73.9 \\

 % \cline{3-11}
 % & \textit{WS3D}~\cite{liu2022weakly} & 45.3~\textbf{\tiny{+9.0}} & 54.6~\textbf{\tiny{+9.6}} & 59.3~\textbf{\tiny{+6.4}} & 62.3~\textbf{\tiny{+7.0}} & 65.7~\textbf{\tiny{+5.8}} & 66.5~\textbf{\tiny{+6.8}} & 67.2~\textbf{\tiny{+6.0}} & 69.5~\textbf{\tiny{+6.9}} & 72.9~\textbf{\tiny{+6.5}} \\
 %  % \cline{3-11}

 %   % \cline{3-11}
 % & \textit{WS3D}~\cite{liu2022weakly} (Pr) & 46.9~\textbf{\tiny{+10.6}} \cellcolor{orange!37} & 56.9~\textbf{\tiny{+11.9}} \cellcolor{orange!37} &  \cellcolor{orange!37} 65.6~\textbf{\tiny{+12.7}} & 67.8~\textbf{\tiny{+14.0}} \cellcolor{orange!37} & 65.7~\textbf{\tiny{+5.8}} \cellcolor{orange!37} & 67.8~\textbf{\tiny{+7.5}} \cellcolor{orange!37} & 72.3~\textbf{\tiny{+11.2}} \cellcolor{orange!37} & 73.9~\textbf{\tiny{+12.7}} \cellcolor{orange!37} & 77.8~\textbf{\tiny{+11.4}} \cellcolor{orange!37} \\
 %  \cline{3-11}
 
  &  \textit{WS3D++} & 48.6~\textbf{\tiny{+12.3}}  & 57.7~\textbf{\tiny{+12.7}} & 61.2~\textbf{\tiny{+8.3}}  & 66.9~\textbf{\tiny{+11.6}} & 70.6~\textbf{\tiny{+10.7}} & 71.1~\textbf{\tiny{+10.8}} & 72.6~\textbf{\tiny{+11.4}} & 75.3~\textbf{\tiny{+12.7}} & 82.0~\textbf{\tiny{+15.6}} \\

 &  \textit{WS3D++} (Pr) & 55.8~\textbf{\tiny{+19.5}} \cellcolor{orange!37} & 64.7~\textbf{\tiny{+19.7}} \cellcolor{orange!37} & 68.2~\textbf{\tiny{+15.3}} \cellcolor{orange!37} & 68.8~\textbf{\tiny{+15.0}} \cellcolor{orange!37} & 75.8~\textbf{\tiny{+15.9}}  \cellcolor{orange!37}& 75.5~\textbf{\tiny{+15.2}} \cellcolor{orange!37} & 75.9~\textbf{\tiny{+14.7}} \cellcolor{orange!37} & 77.8~\textbf{\tiny{+15.2}} \cellcolor{orange!37} & \cellcolor{orange!37} 82.9~\textbf{\tiny{+16.5}} \\

 & Merely Pre-training Stage & 45.2 & 52.8 & 58.7 & 63.8 & 68.6 & 69.8 &  71.2 & 72.6 & 73.9 \\
 
\hline
%  & $\uparrow$ &   \textbf{} 3 \cline\textbf{+9.0/+7.1} & \textbf{+9.6/+1.6} & \textbf{+6.4/+1.6} & \textbf{+7.0/+2.1} & \textbf{+5.8/+2.2} & \textbf{+6.2/+2.6} & \textbf{+6.0/+2.3} & \textbf{+6.9/+4.5} & \textbf{+6.5/+4.1}\\ 33.5
% \hline

\multirow{11}{*}{SemanticKITTI} & Sup-only-GPC--Baseline & 28.6  & 34.8 & 43.9 & 47.9 & 53.8 & 55.1 & 55.4 & 57.4 &  65.0 \\

& Sup-only-GPC~(Pr) & 36.3 & 41.8 & 50.3 & 55.6 & 57.6 & 63.8 & 60.8 & 59.8 & 67.8 \\

 &  \multirow{1}{*}{\textbf{GPC}~\cite{jiang2021guided}} &  34.7 & 41.8 & 49.9 & 53.1 & 58.8 & 59.1 & 59.4 & 59.9 & 65.8 \\

 &  \multirow{1}{*}{\textbf{GPC}}~\cite{jiang2021guided} (Pr) & 36.9 & 45.6 & 53.8 & 55.8 & 59.9 & 63.9 & 65.6 & 64.8 & 69.5 \\

 & LESS~\cite{liu2022less} & 37.1 & 42.5 & 50.5 &  53.9 & 59.5 & 59.6 & 60.5 & 63.5 &  66.9 \\

 & LESS~\cite{liu2022less}~(Pr) & 39.6 & 46.7 & 53.9 & 56.8 & 63.9 & 65.7 & 65.3 & 68.7 &  68.8 \\
 
% \textbf{} \cellcolor{orange!37}
  &  Mix3D~\cite{nekrasov2021mix3d} & 37.7 & 42.9 & 50.8 & 54.1 & 59.9 & 60.9 & 61.8 & 61.3 & 68.8 \\
  &  Mix3D~\cite{nekrasov2021mix3d}~(Pr) & 41.8 & 46.6 & 53.9 & 58.6 & 64.7 & 65.6 & 66.6 & 65.9 & 73.7 \\
  
  % \cline{3-11}
  % & \multirow{1}{*}{\textit{WS3D}}~\cite{liu2022weakly} & 38.9~\textbf{\tiny{+10.3}} & 43.7~\textbf{\tiny{+8.9}} & 52.3~\textbf{\tiny{+8.4}} & 55.5~\textbf{\tiny{+7.6}} & 61.4~\textbf{\tiny{+7.6}} & 61.8~\textbf{\tiny{+6.7}} & 62.1~\textbf{\tiny{+6.7}} &63.2~\textbf{\tiny{+5.8}} &  70.8~\textbf{\tiny{+5.8}} \\

  %   % \cline{3-11}
  % & \multirow{1}{*}{\textit{WS3D}}~\cite{liu2022weakly}~(Pr) & 45.8~\textbf{\tiny{+17.2}} \cellcolor{orange!37} & 46.9~\textbf{\tiny{+12.1}} \cellcolor{orange!37} & 57.6~\textbf{\tiny{+13.7}} \cellcolor{orange!37} & \cellcolor{orange!37} 67.5~\textbf{\tiny{+19.6}}  & 73.4~\textbf{\tiny{+19.6}} \cellcolor{orange!37} & 71.8~\textbf{\tiny{+16.7}} \cellcolor{orange!37} & 73.1~\textbf{\tiny{+17.7}} \cellcolor{orange!37} &72.3~\textbf{\tiny{+14.9}}  \cellcolor{orange!37}&  76.9~\textbf{\tiny{+11.9}} \cellcolor{orange!37} \\

  \cline{3-11}
  &  \textit{WS3D++} & 46.8~\textbf{\tiny{+18.2}}  & 48.6~\textbf{\tiny{+13.8}} & 55.2~\textbf{\tiny{+11.3}} & 62.8~\textbf{\tiny{+14.9}} & 65.9~\textbf{\tiny{+12.1}} & 67.9~\textbf{\tiny{+12.8}} & 68.6~\textbf{\tiny{+13.2}} & 70.9~\textbf{\tiny{+13.5}} & 76.8~\textbf{\tiny{+11.8}} \\

% +18.2
 &  \textit{WS3D++} (Pr) &   \cellcolor{orange!37} 51.6~\textbf{\tiny{+33.0}} & \cellcolor{orange!37} 59.3~\textbf{\tiny{+24.5}} & \cellcolor{orange!37} 68.7~\textbf{\tiny{+24.8}} & \cellcolor{orange!37} 69.8~\textbf{\tiny{+21.9}} & \cellcolor{orange!37} 72.9~\textbf{\tiny{+19.1}} & \cellcolor{orange!37} 74.7~\textbf{\tiny{+19.6}} & \cellcolor{orange!37} 76.3~\textbf{\tiny{+20.9}} & \cellcolor{orange!37} 76.6~\textbf{\tiny{+19.2}} & \cellcolor{orange!37} 83.9~\textbf{\tiny{+18.9}} \\

 & Merely Pre-training Stage & 42.9 & 46.6 & 51.7 & 61.7 & 63.8 & 67.8 &  69.6 & 71.8 & 75.8 \\
\bottomrule
\end{tabular}}
\label{table_sem_1}
\end{center}
 \vspace{-1mm}
\end{table*}

\begin{figure*}[t!]
\centering
\includegraphics[width=\linewidth]{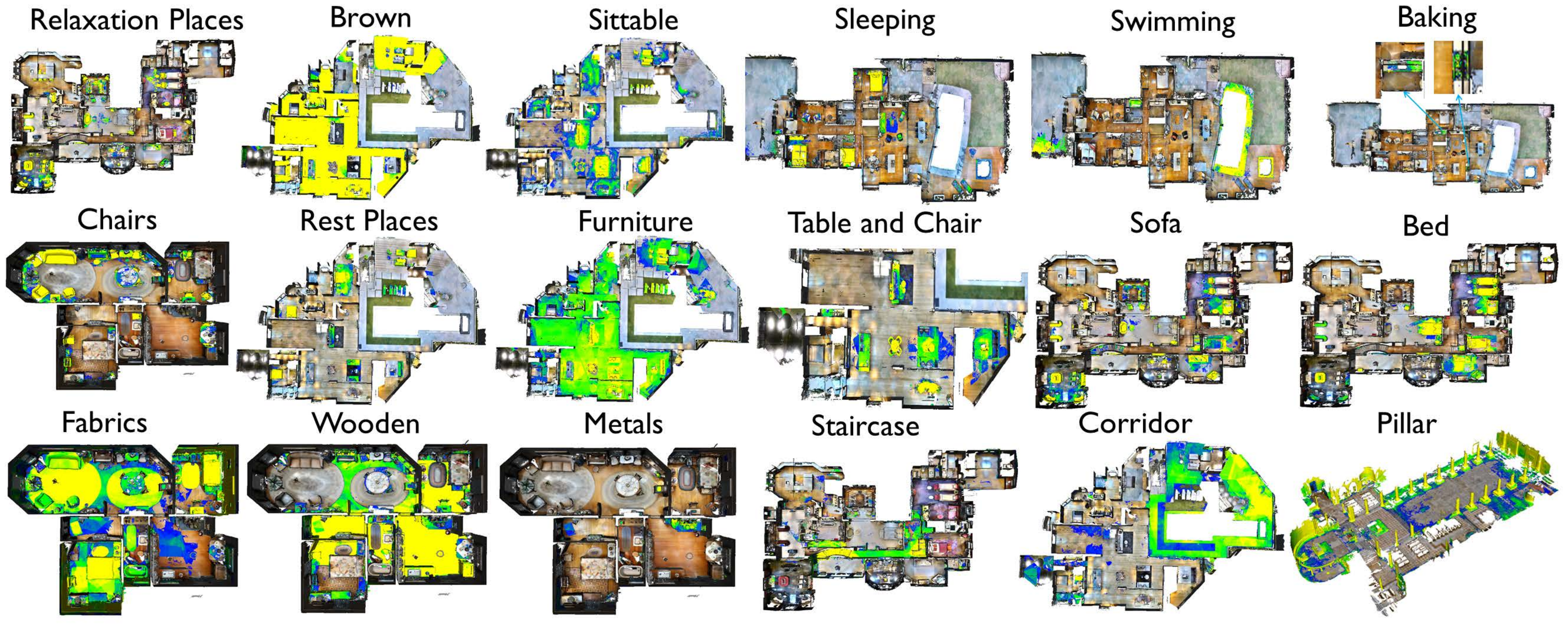}
% \includegraphics[scale=0.15] CLIP_Visualize_Pred_Sem_KITTI.png CLIP_Visualize_Final_Pred.png
% {fig/CLIP_Visualize_Final_Pred.png} CLIP_Visualize_Pred.png s Segmentation result visualizations and comparisons with CLIP prompts for the outdoor KITTI benchmark. % in The s  % \vspace{-4.89mm}width=\linewidth\vspace{+1.89mm} \vspace{-1.96mm
\caption{The final 3D scene-level activations results based on the language query of \textit{WS3D++}. Better zoom in for details.}
\label{fig_scene_act}
\end{figure*}

\begin{figure*}[t]
\centering
\includegraphics[width=\linewidth]{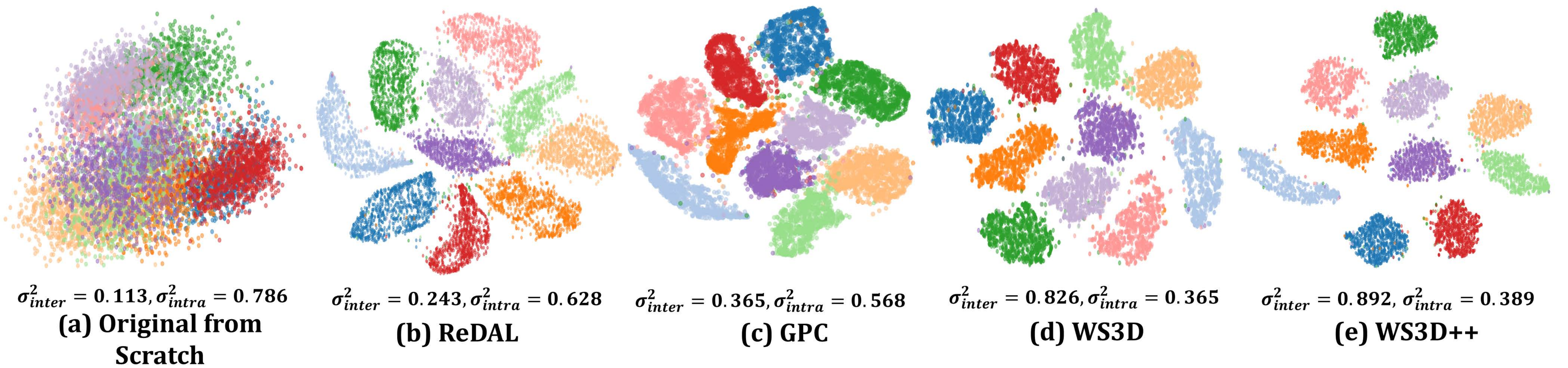}
\caption{\textbf{t-SNE} visualization in semantic segmentation of the proposed \textit{WS3D} under the 5\% labeling percentage on S3DIS validation set. The diverse feature embeddings are indicated by different colors and are normalized into $[-1, 1]$ for better visualization. It has been validated that more discriminative features can be acquired for diverse semantic classes with our proposed unsupervised region-aware fine-tuning strategy (demonstrated by \textit{WS3D}) and our proposed hierarchical vision-language knowledge associated and distilled pre-training (demonstrated by \textit{WS3D++}).}
\label{fig_tsne}
\vspace{-2mm}
\end{figure*}
% \vspace{-1mm} train is to \vspace{-1.6 mm} \vspace{-5mm}

\vspace{-2.1mm}
\noindent
\textbf{Implementation Details.} For the task of semantic segmentation, we fine-tune the network for 500 epochs on a single NVIDIA 1080Ti GPU with a batch size of 16 during training. The initial learning rate is set to 1$\times$10$^{-3}$ and is multiplied with 0.2 every 50 epochs. We implement it in \textit{PyTorch} and optimize it with \textit{Adam} optimizer~\cite{kingma2014adam}. We set the hyperparameter $\gamma$ as 0.8 to ensure that merely highly confident prediction can be used for network optimization. $\epsilon$ is set to 0.5. We empirically choose $\alpha=\beta=1$, while $\lambda_u=0.1$. For instance segmentation, we train the network for 580 epochs on a single NVIDIA 1080Ti GPU with a batch size of 8 during training. The other settings are the same as the semantic segmentation task.

\vspace{+1.2 mm}
% \vspace{-1.1 mm} \vspace{-1.1 mm}

\subsection{Data-efficient 3D Semantic Segmentation}

% \vspace{-0.6mm}

 % 6 in terms of mIoU various the  that compared with Sup-only-GPC models, \textbf{}

\textbf{Overall Experimental Results.} For the semantic segmentation, we have tested \textit{WS3D++} on versatile indoor and outdoor benchmarks, including ScanNet~\cite{dai2017scannet}, S3DIS~\cite{armeni20163d}, and SemanticKITTI~\cite{behley2019semantickitti}. We have done extensive experiments with limited labeled data, \textit{e.g.}, only \{1\%, 5\%, 10\%, 15\%, 20\%, 25\%, 30\%, 40\%, 100\%\} data in training set are available as labeled data. The qualitative results are shown in Fig.~\ref{fig_sem_skitti}. In the meanwhile, the quantitative semantic segmentation performance is summarized in Table~\ref{table_sem_1}. As mentioned, we have used the voxel-based method SparseConv~\cite{graham20183d} as the backbone. Our WSL model significantly surpasses the supervised-only model in \textbf{GPC} that is merely trained with labeled data, showing that our WSL can effectively make use of the unlabeled data to enhance the feature discrimination capacity of the model. Also, it can be observed the increment of performance is more obvious when the unlabeled data percentage is larger. For example, the performance increase on SemanticKITTI is 10.3\% for the 1\% labeling percentage, 5.8\% for the 40\% labeling percentage, and 1.9\% for the 100\% labeling percentage. This can be possibly explained by the fact that for more unlabeled data, our proposed \textit{WS3D++} can extract more meaningful semantic information from the unlabeled data based on our boundary-guided energy-based loss and confidence-guided region-level contrastive learning design. In addition, compared with current SOTA \textbf{GPC}, our proposed \textit{WS3D++} also achieves consistently better results in semantic segmentation performance, especially when faced with very limited label circumstances (e.g. 1\% labeling points).
In that case, \textit{WS3D++} outperforms \textbf{GPC} by 3.3\%, 7.1\%, and 4.2\% for ScanNet, S3DIS, and SemanticKITTI, respectively. Fig.~\ref{fig_sem_skitti} shows that we can provide comparable performance compared with fully supervised SOTAs BAAF-Net~\cite{qiu2021semantic} and Cylinder3D~\cite{zhu2021cylindrical} on SemanticKITTI with 5\% labels. As shown in Table~\ref{table_sem_1}, the performance of our enhanced approach~\textit{WS3D++} has remarkably increased performance compared with \textit{WS3D++} and previous SOTAs, demonstrating the effectiveness of our proposed vision-language knowledge-associated pre-training. 
% \vspace{-2.2mm} \textcolor{blue}{ }  [1]

We have reported the experimental results about semantic boundary prediction as demonstrated in Table 1 of our revised manuscript, it is demonstrated that our proposed approach can provide a very high-quality label at a very low-label regime, and the boundary prediction average precision can be kept at more than 50\% (50.9\% at least). The phenomenon has also been found at previous research including SQN, Lasermix, and our Weaklab3DNet~\cite{liu2022weaklabel3d}. The reason behind is that: As point clouds are essentially samples of the 3D world, the distribution of points in a very close local neighborhood is comparatively homogeneous, revealing strong semantic similarity/homogeneity. Moreover, our proposed weakly supervised approach can be regarded as an amplification of those rather sparse supervision signals, which largely facilitates ultimate semantic boundary prediction. As we have demonstrated in Table I, the average precision (AP) of boundary prediction can still be maintained at a relatively high value (more than 50.9\%). Our proposed approach can realize promising performance on outdoor benchmark SemanticKITTI and indoor benchmarks including S3DIS and ScanNet, which demonstrates the superior effectiveness of our proposed weakly supervised semantic boundary prediction.
% \vspace{-2mm}
% \vspace{2mm}

% We have tested extensively on diverse benchmarks and s \subsection{\vspace{-1.0mm} 5

 % apparent increment ++ our proposed  

% improved 

\begin{figure*}[t]

\centering
\includegraphics[width=\linewidth]{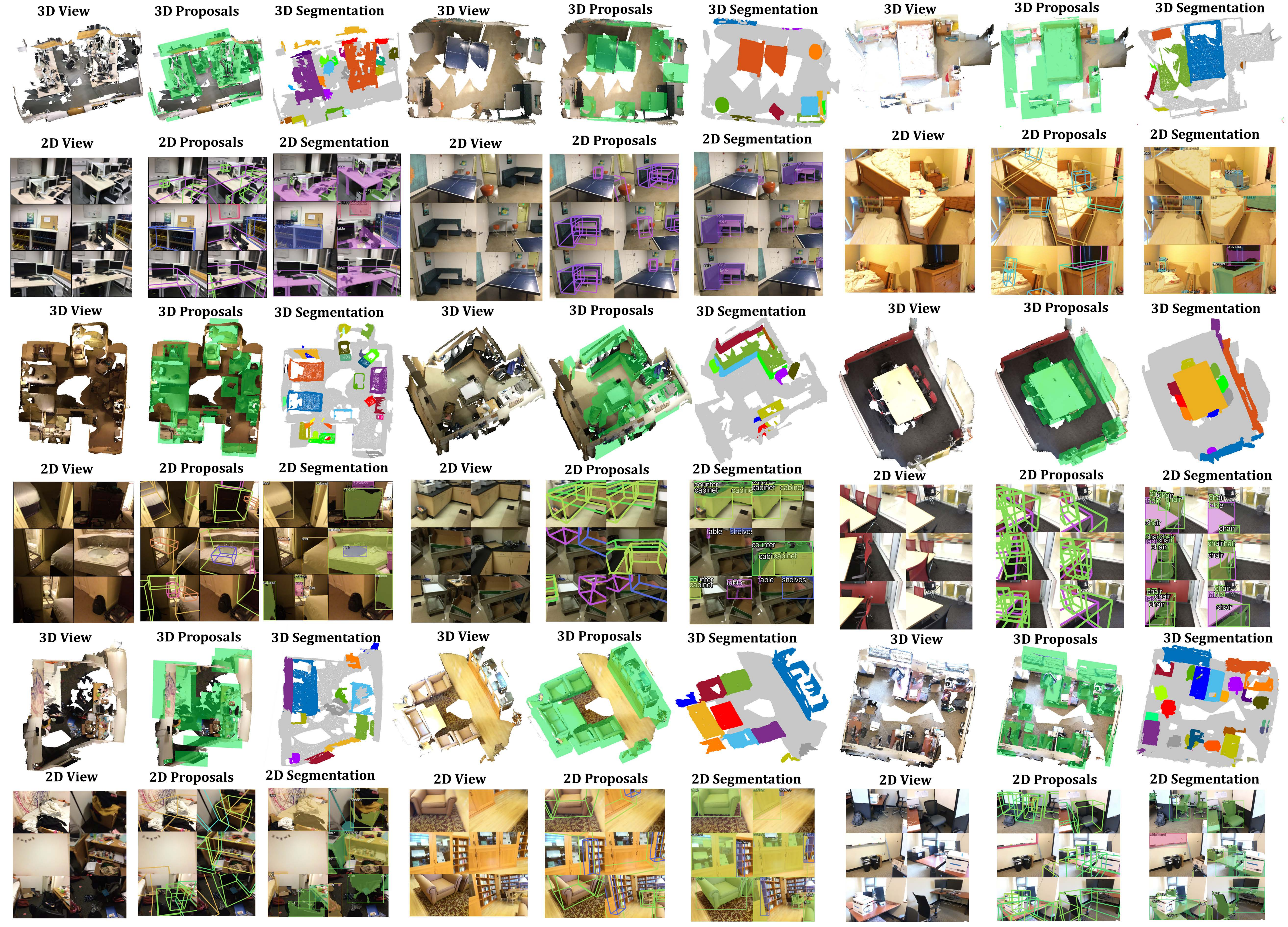}
\caption{The \textit{WS3D++} corresponding detection and segmentation projections on six rendered 2D views through our proposed multi-view rendering approach. It can be demonstrated that our proposed rendering has established an explicit modality association between the final 2D views and 3D views. Also, by combining 2D and 3D region proposals, more complete and apparent object-level information can be clearly captured both from 2D views and 3D views. Best zoom-in for viewing.}
\label{fig_views}
\vspace{-1.628mm}
\end{figure*}
%%%%%%%%%%%%Table 2, 3or sometimes even better 8 38     \vspace{-1.58mm} \vspace{+1.628mm} \vspace{-1.628mm}

 \begin{table}[t]
    \centering    \caption{Comparison of experimental results on 20\% and fully labeled case for the task of inductive and transductive learning for our proposed \textit{WS3D++}, respectively. In transductive learning, the test set is also utilized for network training. We test the task of semantic segmentation on ScanNet, S3DIS, and SemanticKITTI with the evaluation metric of mIoU(\%).}
    % \vspace{-1.38mm}
\label{table_transduct}
\resizebox{\linewidth}{!}{
    \begin{tabular}{l|ccc|ccc}
    \toprule
    \multirow{2}{*}{Datasets} & \multicolumn{3}{c|}{20\% label} &\multicolumn{3}{c}{100\% label}\cr
    \cline{2-7}
     &   & \\[-1em]
    & {Base} & Induct. & Transduct.&{Base} & Induct. & Transduct. \\
    \hline
    % \multirow{1}{*}{ScanNetv2-\textit{WS3D}} & 64.0 & 68.5 & 71.4 & 72.9 & 76.9 & 77.6  \\

    % \multirow{1}{*}{ScanNetv2-\textit{WS3D}} (Pr) & 69.6 & 66.8 & 86.3 & 82.6 & 83.8 & 88.9 \\
    
    \multirow{1}{*}{ScanNetv2-\textit{WS3D++}} & 69.9 & 71.8 & 85.8 & 78.9 & 81.7 & 87.8  \\
    \multirow{1}{*}{ScanNetv2-\textit{WS3D++}} (Pr) & 75.3 & 79.8 & 87.8 & 85.6 & 88.8 & 91.8 \\
    % \multirow{1}{*}{S3DIS Area5 Val.}  & 59.9 & 65.7 & 66.6 & 66.4 & 72.9 & 73.5 \\
    % \multirow{1}{*}{S3DIS Area5 Val.} (Pr)  & 68.6 & 78.6 & 78.7 & 79.6 & 86.7 & 85.9 \\
    % \multirow{1}{*}{S3DIS \textit{WS3D}}  & 61.3 & 67.6 & 68.6 & 68.2 & 74.8 & 75.8 \\
    % \multirow{1}{*}{S3DIS \textit{WS3D}} (Pr)  & 72.7 & 73.6 & 82.8 & 78.6 & 83.9 & 87.9 \\
    \multirow{1}{*}{S3DIS \textit{WS3D++}}  & 62.8 & 70.6 & 75.8 & 69.7 & 82.0 & 86.8 \\
    \multirow{1}{*}{S3DIS \textit{WS3D++}} (Pr) & 68.7 & 75.8 & 78.7 & 73.8 & 82.9 & 87.3 \\   
    % \multirow{1}{*}{Semantic KITTI Val.} & 53.8 & 61.4 & 64.5 & 65.0 & 66.9 & 68.2\\
    % \multirow{1}{*}{Semantic KITTI Val.} (Pr)  & 58.9 & 68.7 & 69.3 & 69.5 & 73.9 & 69.7 \\    
    \multirow{1}{*}{Semantic KITTI Val. \textit{WS3D++}}  & 58.9 & 65.9 & 79.3 & 74.3 & 76.8 & 88.8\\
   \multirow{1}{*}{Semantic KITTI Val. \textit{WS3D++}} (Pr)  & 62.8 & 72.9 & 69.1 & 68.6 & 83.9 & 85.5 \\    
    % \multirow{1}{*}{Semantic KITTI Test.}  & 55.7 & 62.5 & 63.6 & 65.4 & 68.1 & 71.3 \\

    % \multirow{1}{*}{Semantic KITTI Test.} (Pr) & 58.5 & 65.8 & 65.9 & 69.2 & 73.6 & 76.2 \\
    
    \multirow{1}{*}{Semantic KITTI Test. \textit{WS3D++}} & 66.7 & 71.8 & 72.6 & 71.6 & 77.9 & 79.6 \\
    \multirow{1}{*}{Semantic KITTI Test. \textit{WS3D++}} (Pr) & 70.8 & 75.9 & 78.8 & 79.1 & 84.3 & 86.5 \\
    \bottomrule
    \end{tabular}}

\end{table}
    % \vspace{-3.1 mm}    \vspace{-3.1 mm}
% \textbf{47.91} \textbf{47.91} \% \textbf{47.91} ($\downarrow$ ) ($\downarrow$  \%) ( $\downarrow$ \%) ($\downarrow$ \textbf{198.43} \%)     \vspace{-2.1 mm}

 % Val. Area5 Val. ++      \vspace{-1.58mm} table_transduct_2 6     \multirow{6}{*}{SUN RGB-D~\cite{song2015sun, liu2022weaklabel3d}} \textit{WS3D-Open} is examined to validate the effect of hierarchical feature aligned pre-training in improving data efficiency. We have also tested the ablated approach \textit{WS3D-Open} which abandons our feature-aligned pre-training term $\mathcal{L}^{total}_{Ctr}$ and directly uses the CLIP~\cite{radford2021learning} feature encoder for the knowledge distillation loss term with $\mathcal{L}^{Dist}_{\mathcal{KL}}$. It can be also demonstrated that our proposed pre-training framework can boost the scene parsing performance with our proposed effective hirarchical feature aligned pre-training designs.

\begin{table*}[t]
    \centering    \caption{Comparison of current State-of-the-art (SOTA) approaches in the limited reconstruction case for the \textbf{3D object detection} tasks with different ratios of labeled data. The mean average precision (mAP) given also with  the mean $\pm$ {standard deviation} in three runs of diverse random splits are reported. Baseline is the circumstance which is trained with merely labelled data. It can be demonstrated our proposed pre-training approach have improved the performance of 3D object detection by a considerable margin.  }  \renewcommand\arraystretch{1.32}\label{table_transduct_2}    \resizebox{\linewidth}{!}{
    \begin{tabular}{c|l|cc|cc|cc|cc}
    \toprule
    \multirow{2}{*}{Datasets} & \multirow{2}{*}{Models} & \multicolumn{2}{c|}{5\% } &\multicolumn{2}{c|}{10\% } &\multicolumn{2}{c}{20\%}&\multicolumn{2}{c}{100\%}\cr
    \cline{3-10}
     &   & \\[-1em] &
    &  Induct. & Transduct.&Induct. & Transduct. & Induct. & Transduct. & Induct. & Transduct. \\
    \hline
    \multirow{11}{*}{SUN RGB-D~\cite{song2015sun, liu2022weaklabel3d}} & Baseline & 29.9 $\pm$ 1.5 &  33.5 $\pm$ 0.8 & 34.4 $\pm$ 1.1 & 40.7 $\pm$ 0.9 & 41.1 $\pm$ 0.3 & 47.5 $\pm$ 0.5 & 51.7 $\pm$ 0.9 & 53.8 $\pm$ 0.7  \\

    % & Baseline (Pr) & 33.9 $\pm$ 1.6 & 38.6 $\pm$ 1.7 & 39.6 $\pm$ 5.6 & 45.8 $\pm$ 2.6 & 47.8 $\pm$ 0.9 & 56.8 $\pm$ 0.6 & 49.2 $\pm$ 0.4 & 58.7 $\pm$ 0.3 \\
     
    & 
     
     Merely the Pre-training Stage &  
     35.8 $\pm$ 1.1 & 38.2 $\pm$ 1.1 & 
     39.3 $\pm$ 0.7 & 
     45.1 $\pm$ 0.7 & 
     46.1 $\pm$ 0.9 &
     56.9 $\pm$ 0.7 &  
     51.9 $\pm$ 1.1 & 
     65.8 $\pm$ 1.5 \\
     &SESS~\cite{zhao2020sess} & 34.2 $\pm$ 2.0 & 38.1 $\pm$ 0.7  & 42.9 $\pm$ 0.8 & 45.3$\pm$ 0.9 & 47.9 $\pm$ 0.4 & 51.6 $\pm$ 0.2 & 50.7 $\pm$ 0.5 & 53.7 $\pm$ 0.5 \\
     
     &SESS (Pr)~\cite{zhao2020sess} & 42.8 $\pm$ 3.2 & 42.6 $\pm$ 0.7 & 48.2 $\pm$ 0.8 & 53.8 $\pm$ 0.9 & 50.5 $\pm$ 0.6 & 56.3 $\pm$ 0.7 & 55.2 $\pm$ 0.3 & 63.9 $\pm$ 0.6 \\

   &3D-IOUMatch~\cite{wang20213dioumatch} & 39.1 $\pm$ 1.9 & 46.3 $\pm$ 0.7 & 45.5 $\pm$ 1.5 & 53.5 $\pm$ 0.3 & 51.6 $\pm$ 0.5 & 55.6 $\pm$ 0.8 & 55.6 $\pm$ 0.3 & 62.3 $\pm$ 0.7 \\

    &3D-IOUMatch (Pr)~\cite{wang20213dioumatch} & 43.6 $\pm$ 3.2 & 48.5 $\pm$ 0.9 & 48.5 $\pm$ 1.8 & 56.9 $\pm$ 0.5 & 59.8 $\pm$ 0.7 & 56.8 $\pm$ 0.8 & 65.7 $\pm$ 0.5 & 69.1 $\pm$ 0.7 \\
      % &SPD~\cite{xie2022spd} & 38.5 $\pm$ 0.7 & 44.3 $\pm$ 0.8 & 46.0 $\pm$ 1.0 & 51.7 $\pm$ 0.9 & 49.6 $\pm$ 0.5 & 54.5 $\pm$ 0.9 \\
      % & \cellcolor[RGB]{232,232,232} \textit{WS3D} (Ours) & \cellcolor[RGB]{232,232,232} 46.9 $\pm$ 0.8 & \cellcolor[RGB]{232,232,232} 49.5 $\pm$ 0.7 & \cellcolor[RGB]{232,232,232} 47.8 $\pm$ 0.8 & \cellcolor[RGB]{232,232,232} 55.7 $\pm$ 0.8 & \cellcolor[RGB]{232,232,232} 55.6 $\pm$ 0.7 & \cellcolor[RGB]{232,232,232} 57.9 $\pm$ 1.2 \\ \cellcolor[RGB]{232,232,232} \cellcolor[RGB]{232,232,232} \cellcolor[RGB]{232,232,232}

    &SPD~\cite{xie2022spd} & 38.5 $\pm$ 0.7 & 44.3 $\pm$ 0.8 & 46.0 $\pm$ 1.0 & 51.7 $\pm$ 0.9 & 60.9 $\pm$ 0.5 & 56.9 $\pm$ 0.9 & 67.9 $\pm$ 0.7 & 73.9 $\pm$ 0.8 \\

    &SPD~\cite{xie2022spd} (Pr) & 43.6 $\pm$ 1.2 & 47.8 $\pm$ 1.7 & 49.6 $\pm$ 0.8 & 53.8 $\pm$ 1.1 & 58.6 $\pm$ 1.6 & 68.9 $\pm$ 2.5 & 69.2 $\pm$ 0.9 & 83.8 $\pm$ 0.5 \\

    % &  \textit{WS3D} (Ours) &  46.9 $\pm$ 0.8 &  49.5 $\pm$ 0.7 &  47.8 $\pm$ 0.8 &  55.7 $\pm$ 0.8 &  55.6 $\pm$ 0.7 &  57.9 $\pm$ 1.2 \\

     %   & \cellcolor{orange!37} \textit{WS3D} (Ours) (Pr) & \cellcolor{orange!37} 49.7 $\pm$ 0.9 & \cellcolor{orange!37} 52.9 $\pm$ 1.2 & \cellcolor{orange!37} 51.8 $\pm$ 1.7 & \cellcolor{orange!37} 67.8 $\pm$ 1.3 & \cellcolor{orange!37} 59.6 $\pm$ 1.5 & \cellcolor{orange!37} 66.7 $\pm$ 1.2 \\
      
      % & \cellcolor[RGB]{232,232,232} \textit{WS3D-Open} (Ours) & \cellcolor[RGB]{232,232,232} 47.8 $\pm$ 0.8 & \cellcolor[RGB]{232,232,232} 50.3 $\pm$ 0.6 & \cellcolor[RGB]{232,232,232} 49.3 $\pm$ 0.7 & \cellcolor[RGB]{232,232,232} 58.2 $\pm$ 0.7 & \cellcolor[RGB]{232,232,232} 59.1 $\pm$ 0.8 & \cellcolor[RGB]{232,232,232} 60.3 $\pm$ 1.1 \\ \cellcolor[RGB]{232,232,232} \cellcolor[RGB]{232,232,232} \cellcolor[RGB]{232,232,232} \cellcolor[RGB]{232,232,232} \cellcolor[RGB]{232,232,232} \cellcolor[RGB]{232,232,232} \cellcolor[RGB]{232,232,232}
      &  \textit{WS3D++} (Ours) &  52.7 $\pm$ 0.7 &  54.5 $\pm$ 0.8 &  53.7 $\pm$ 0.9 & 63.8 $\pm$ 0.8 &  69.8 $\pm$ 0.7 & 72.9 $\pm$ 1.1 & 68.6 $\pm$ 0.9 & 82.8 $\pm$ 0.5 \\

      & \cellcolor{orange!37} \textit{WS3D++} (Ours) (Pr) & \cellcolor{orange!37} 56.9 $\pm$ 0.8 & \cellcolor{orange!37} 59.7 $\pm$ 1.5 & \cellcolor{orange!37} 58.7 $\pm$ 1.8 & \cellcolor{orange!37} 68.9 $\pm$ 1.3 & \cellcolor{orange!37} 68.6 $\pm$ 1.8 & \cellcolor{orange!37} 73.7 $\pm$ 1.6  &  \cellcolor{orange!37} 72.6 $\pm$ 1.1 & \cellcolor{orange!37} 83.9 $\pm$ 0.9 \\

      \hline
      \multirow{11}{*}{ScanNet-V2~\cite{dai2017scannet}} & Baseline& 27.9 $\pm$ 0.5 & 30.8 $\pm$ 1.5 & 31.1 $\pm$ 0.7 & 33.2 $\pm$ 0.5 & 41.6 $\pm$ 0.9 & 44.5 $\pm$ 1.2 & 45.8 $\pm$ 0.8 & 48.9 $\pm$ 0.9 \\

  % &Baseline (Pr)& 36.8 $\pm$ 0.5 & 35.2 $\pm$ 2.5 & 37.6 $\pm$ 0.8 & 38.1 $\pm$ 0.5 & 45.7 $\pm$ 0.9 & 47.8 $\pm$ 1.2 & 50.6 $\pm$ 0.4 & 54.5 $\pm$ 0.9 \\

    & 
     Merely the Pre-training Stage & 35.2 $\pm$ 1.5 &  32.8 $\pm$ 1.1 & 
     35.6 $\pm$ 0.8 & 
     36.6 $\pm$ 0.6 & 
     43.9 $\pm$ 1.1 &  
     47.9 $\pm$ 0.8 &  
     49.3 $\pm$ 0.7 & 
     56.8 $\pm$ 0.9 \\
      % Baseline (Pr)& 27.9 $\pm$ 0.5 & 30.8 $\pm$ 1.5 & 31.1 $\pm$ 0.7 & 33.2 $\pm$ 0.5 & 41.6 $\pm$ 0.9 & 44.5 $\pm$ 1.2 \\ \cellcolor[RGB]{232,232,232}
    &SESS~\cite{zhao2020sess} & 32.2 $\pm$ 0.8 & 36.8 $\pm$ 1.1  & 39.7 $\pm$ 0.9 & 44.5 $\pm$ 0.5 & 48.2 $\pm$ 0.4 & 49.6 $\pm$ 0.7 & 55.8 $\pm$ 0.1 & 56.8 $\pm$ 0.7\\
    &SESS~\cite{zhao2020sess} (Pr) & 35.9 $\pm$ 0.9 & 43.9 $\pm$ 1.2  & 43.5 $\pm$ 1.2 & 48.9 $\pm$ 0.5 & 49.6 $\pm$ 0.6 & 53.2 $\pm$ 0.7 & 56.9 $\pm$ 0.6 & 59.6 $\pm$ 0.9 \\
      
    &3D-IOUMatch~\cite{wang20213dioumatch} & 40.0 $\pm$ 0.9 & 46.3 $\pm$ 0.7 & 47.2 $\pm$ 0.4 & 49.6 $\pm$ 0.6 & 52.8 $\pm$ 1.2 & 62.9 $\pm$ 0.8 & 56.9 $\pm$ 1.2 & 63.7 $\pm$ 0.8 \\

    &3D-IOUMatch (Pr)~\cite{wang20213dioumatch} & 45.6 $\pm$ 2.0 & 52.8 $\pm$ 0.8 & 53.3 $\pm$ 1.5 & 55.6 $\pm$ 1.7 & 57.7 $\pm$ 1.2 & 59.6 $\pm$ 0.8 & 61.8 $\pm$ 1.6 & 66.9 $\pm$ 1.3 \\
      
    &SPD~\cite{xie2022spd} & 41.5 $\pm$ 0.5 & 44.3 $\pm$ 0.8 & 43.2 $\pm$ 1.2 & 46.2 $\pm$ 0.5 & 51.9 $\pm$ 0.5 & 55.6 $\pm$ 0.9 & 54.8 $\pm$ 0.5 & 58.6 $\pm$ 0.9 \\

    &SPD~\cite{xie2022spd} (Pr) & 45.2 $\pm$ 0.7 & 48.8 $\pm$ 1.9 & 52.8 $\pm$ 1.6 & 55.2 $\pm$ 0.7 & 53.8 $\pm$ 1.8 & 65.2 $\pm$ 2.5 & 59.9 $\pm$ 0.5 & 68.9 $\pm$ 1.6 \\

   &  \textit{WS3D++} (Ours) & 55.3 $\pm$ 1.1 &  57.6 $\pm$ 1.2 &  59.8 $\pm$ 0.8 &  66.8 $\pm$ 1.2 &  64.6 $\pm$ 0.5 & 72.8 $\pm$ 1.1 &  68.7 $\pm$ 0.5 &  76.6 $\pm$ 1.1 \\

   & \cellcolor{orange!37} \textit{WS3D++} (Ours) (Pr) & \cellcolor{orange!37} 66.6 $\pm$ 1.5 & \cellcolor{orange!37} 68.1 $\pm$ 1.7 & \cellcolor{orange!37} 67.6 $\pm$ 1.6 & \cellcolor{orange!37} 74.5 $\pm$ 1.3 & \cellcolor{orange!37} 69.5 $\pm$ 0.9 & \cellcolor{orange!37} 76.9 $\pm$ 0.9 & \cellcolor{orange!37} 73.6 $\pm$ 0.7 & \cellcolor{orange!37} 77.8 $\pm$ 0.6 \\
      % & \cellcolor[RGB]{232,232,232} \textit{WS3D++} (Ours) & \cellcolor[RGB]{232,232,232} 62.8 $\pm$ 1.1 & \cellcolor[RGB]{232,232,232} 65.8 $\pm$ 1.2 & \cellcolor[RGB]{232,232,232} 63.8 $\pm$ 0.8 & \cellcolor[RGB]{232,232,232} 69.8 $\pm$ 1.2 & \cellcolor[RGB]{232,232,232} 68.7 $\pm$ 0.5 & \cellcolor[RGB]{232,232,232} 73.5 $\pm$ 1.1 \\ \cellcolor{orange!37}  \vspace{5mm} \vspace{-5mm}    \vspace{-1mm} \vspace{-0.2mm}  \vspace{-2.2mm}  
    \bottomrule
    \end{tabular}}  \vspace{-2.2mm}  
\end{table*}

\begin{table*}[h]

\caption{Comparison of the performance of \textbf{instance segmentation}, under various levels of supervision on ScanNet validation set.  'Sup-only-GPC' denotes the model trained with only labeled data. '\textit{WS3D}' denotes the model trained with our proposed methods. In the last row, we have shown the performance increase of \textit{WS3D}. \textit{WS3D-Open} abandons our feature-aligned contrastive pre-training with the contrastive loss term $\mathcal{L}^{total}_{Ctr}$ and directly uses the CLIP~\cite{radford2021learning} feature encoder with the knowledge distillation loss term $\mathcal{L}^{Dist}_{\mathcal{KL}}$. \textit{WS3D-Open} is examined to validate the effect of hierarchical feature-aligned pretraining in improving data efficiency. The results of our proposed approach are highlighted in orange.}
\renewcommand\arraystretch{1.12}

\begin{center}
\scriptsize

\resizebox{\linewidth}{!}{\begin{tabular}{c|l|ccccccccc}
\hline
\multirow{2}{*}{Tested Dataset} & \multirow{2}{*}{Approaches} &\multicolumn{8}{c}{Ins. Seg. Results with the metric of AP@50\%} & \cr \cline{3-11}& &1\% & 5\% & 10\% & 15\% & 20\% & 30\% &35\%& 40\% & 100\%\\
\hline
\multirow{10}{*}{ScanNet} 
&  Merely the Pre-training Stage & 39.7 & 46.7 & 53.8 & 55.6 & 58.6 & 63.5 & 65.2 & 65.7 & 65.8 \\

&  \multirow{1}{*}{Sup-only-GPC--Baseline}~\cite{jiang2021guided} & 10.8 & 33.6 & 42.8 & 45.3 & 48.2 & 49.0 &49.5 & 50.2 & 56.8 \\

&  \multirow{1}{*}{Sup-only-GPC} (Pr)~\cite{jiang2021guided} & 16.8 & 35.9 & 33.8 & 48.9 & 56.6 & 57.9 & 58.8 & 66.8 & 68.9 \\
& \multirow{1}{*}{Mix3D~\cite{nekrasov2021mix3d}} & 12.7 & 34.7 & 43.1 & 45.7 & 48.7 & 49.6 & 50.2 & 51.3 & 57.6 \\

& \multirow{1}{*}{Mix3D (Pr)~\cite{nekrasov2021mix3d}}  & 20.6 & 39.7 & 48.9 & 50.9 & 53.9 & 58.9 & 58.7 & 55.2 & 63.9 \\

& \multirow{1}{*}{GPC}~\cite{jiang2021guided} & 16.9 & 38.6  & 44.9 & 47.2 & 48.6 & 49.7 & 51.2 & 52.0 & 57.7 \\

& \multirow{1}{*}{GPC} (Pr)~\cite{jiang2021guided} & 23.8 & 45.8 & 53.8 & 52.8 & 54.5 & 55.1 & 57.2 & 57.1 & 60.6 \\

& \multirow{1}{*}{SPIB\_Ins}~\cite{liao2021point} & 17.1 & 38.9 & 45.3 & 47.9 & 48.9 & 50.5 & 51.6 & 52.7 & 57.6 \\

& \multirow{1}{*}{SPIB\_Ins} (Pr)~\cite{liao2021point} & 23.7 & 52.8 & 52.9 & 55.3 & 54.8 & 55.8 & 56.9 & 58.8 & 64.8 \\
  & \cellcolor{orange!37} \multirow{1}{*}{\textit{WS3D++} (Ours)} & \cellcolor{orange!37}\textbf{45.9} & \cellcolor{orange!37} \textbf{61.8} & \cellcolor{orange!37}  \textbf{66.8} &  \cellcolor{orange!37}  \textbf{67.7} & \cellcolor{orange!37}  \textbf{65.1} & \cellcolor{orange!37}  \textbf{65.2} & \cellcolor{orange!37}  \textbf{62.7} & \cellcolor{orange!37}  \textbf{68.7} & \cellcolor{orange!37}  \textbf{58.8}
    \\    
    & \cellcolor{orange!37} \multirow{1}{*}{\textit{WS3D++} (Ours)} (Pr) & \cellcolor{orange!37}  \textbf{48.2} & \cellcolor{orange!37} \textbf{72.2} & \cellcolor{orange!37}  \textbf{71.6} &  \cellcolor{orange!37} \textbf{70.5} & \cellcolor{orange!37} \textbf{67.9} & \cellcolor{orange!37}  \textbf{68.7} & \cellcolor{orange!37} \textbf{71.6} & \cellcolor{orange!37} \textbf{75.8} & \cellcolor{orange!37} \textbf{73.9}
    \\
 % & $\uparrow$ & \textbf{+21.7} & \textbf{+12.0} & \textbf{+6.4} & \textbf{+5.8} & \textbf{+3.1}  & \textbf{+2.9} & \textbf{+3.0} & \textbf{+2.8} & \textbf{+1.9} \\ \uparrow $\uparrow$ \textbf{56.8}
% \hline +9.5
\hline
\end{tabular}}
\label{table_ins}
\end{center}
% \vspace{-8mm} WSL-based  \textbf{} \vspace{-3.9mm} 2 5 2 8.6
% \vspace{-5.6mm} % \vspace{-5.6mm}
\vspace{-0.6mm}
\end{table*}

\subsection{Data-Efficient 3D Instance Segmentation}
\vspace{-1.0mm}
As our method can be integrated seamlessly into various network backbones and applied to different highly-level understanding tasks, we have also integrated our method with Point-Group \cite{jiang2020pointgroup} for the instance segmentation on ScanNet with results shown in Table \ref{table_ins}. Notice that the performance increase is 21.7\% when merely 1\% data is labeled compared with the sup-only case. It further demonstrates that our proposed approaches for the unsupervised branch have effectively exploited the unlabeled data to improve the feature learning capacity of the model. Our proposed \textit{WS3D} and \textit{WS3D++} both provide explicit boundary guidance for separating diverse kinds of semantic classes, and the instance segmentation performance with very limited labeling percentage is comparable to those fully supervised counterparts. 
\vspace{-1mm}
% \vspace{-1mm}\vspace{-6mm}\vspace{-1mm}23\vspace{-23mm}\vspace{-3mm}\vspace{-0.9666mm}
% \textit{} \scalebox{1.3}{ as the task of  the  I \vspace{-1.2666mm} 1 3 2 \vspace{-1.6mm} \vspace{-1.6mm} 7 2 we evaluate

\subsection{Data-Efficient 3D Object Detection}
\vspace{-0.18mm}
For the data-efficient 3D object detection, following our previous work~\cite{liu2022rm3d}, we extensively evaluate current approaches extensively on SUN RGB-D~\cite{song2015sun} and ScanNet~\cite{dai2017scannet} benchmarks for 3D object detection tasks with the strong 3D object detection backbone VoteNet~\cite{qi2019deep}. It can be demonstrated the open-vocabulary designs can to some extent boost the 3D object detection performance, which demonstrate the generalization capacity of our fundation model. As shown in Table~\ref{table_transduct_2}, we have also tested the ablated approach \textit{WS3D-Open} which abandons our feature-aligned pre-training term $\mathcal{L}^{total}_{Ctr}$ and directly uses the CLIP~\cite{radford2021learning} feature encoder for the knowledge distillation loss term with $\mathcal{L}^{Dist}_{\mathcal{KL}}$. It can be demonstrated that the performance degradation can be observed when comparing \textit{WS3D-Open} with \textit{WS3D++}, which demonstrate the effectiveness of our proposed hierarchical feature aligned pre-training in improving the data efficiency in downstream scene parsing.

\subsection{ Qualitative and Quantitative Results of the Open-world 3D Recognition Approaches }
\vspace{-0.88mm}
In this Subsection, we further evaluate the performance of the open-world recognition capacity of our proposed approach. The results of open-world recognition are shown in  Table~\ref{table_open_world_scannet}. We have also compared our work with the previous approach PLA~\cite{ding2023pla} in establishing the sufficient point-language associations for the open-world robot learning. The results demonstrate that our proposed approach has superior performance in open-world recognition. We directly use the settings in the PLA~\cite{ding2023pla} and split the categories on ScanNet~\cite{dai2017scannet} and Nuscene~\cite{caesar2020nuscenes} into base and novel categories.  It can also be validated that \textit{WS3D-Open}, which abandons our feature-aligned pre-training and directly use the CLIP~\cite{radford2021learning} feature encoder, provides slightly inferior performance compared with \textit{WS3D++}, validating the effectiveness of our language-3D matching strategy designs. \textit{WS3D++} exhibits superior performance in terms of the open-vocabulary few-shot learning for diverse partitioning of original and novel classes. The open-world recognition results are shown in Figure~\ref{fig_CLIP_Scannet} and Figure~\ref{fig_CLIP_KITTI}. It can be demonstrated that better foreground object awareness can be effectively capture by our proposed \textit{WS3D++} compared with PLA~\cite{ding2023pla}, with superior segmentation performance guided by the textual prompts. The superior open-world recognition performance can be achieved while conducting open-world learning in diverse spliting of based and novel classes, including B15/N4, B12/N7, B10/N9 for the ScanNet~\cite{dai2017scannet} as well as  B12/N3 and B10/N5 for the NuScenes~\cite{caesar2020nuscenes}. It demonstrates robustness of our proposed approach. Also, as demonstrated in Figure~\ref{fig_scene_act}, the \textit{WS3D++} language driven-3D scene segmentation results are very precise as shown qualitatively, which is corresponding to the object queried by the language, and it demonstrates that the inference can be done based on the object, material, properties, affordance, room type, etc. It demonstrates that our proposed \textit{WS3D++} can enable the scene-level object recognition based on the semantic language queries. As further shown in Figure~\ref{fig_views}, through our effective rendering techniques, which establish the explicit 2D-3D association, the aligned representation of 2D-3D-language co-embeddings can be learned and the object information can also be enhanced through finding the similarity among diverse views through contrastive learning approaches. Also, by combining 2D and 3D region proposals, more complete and apparent object-level information can be clearly captured both from 2D views and 3D views.
It turns out that in the first place, while initializing other various weakly supervised approaches, our proposed approach can realize consistent improvement on the final performance of weakly supervised scene parsing, which demonstrates the superior generalization capacity benefiting from our designed pre-training of our proposed WS3D++ framework. In the second pace, the performance of our proposed WS3D++ is comparatively superior compared with the existing weakly supervised comparative approaches listed above.

% \vspace{-0mm} 
 % a 11 11 9 \vspace{-9mm} objectness
% \vspace{-2.8mm} \textit{WS3D-Open} which abandon our feature-aligned pre-training and directly use the CLIP~\cite{radford2021learning} feature encoder
% \textit{WS3D++} As demonstrated in 6
% \vspace{-1.2mm}
\noindent
\subsection{Instance Discrimination Capacity}
% \vspace{-0.1mm}
We show t-SNE visualizations of the learned latent feature representations for various semantic classes in Fig.~\ref{fig_tsne}. The case study task is the semantic segmentation on the S3DIS dataset with a supervision level of 5\%. It is demonstrated that more distinctive and better separated point-wise feature embeddings are provided by our proposed unsupervised region-level contrastive learning, which can be attributed to its strong instance discrimination capacity. And more separated feature space can be provided and maintained with our proposed \textit{WS3D++} compared to \textit{WS3D} and GPC. This strong instance discrimination capacity can be explained by more discriminative feature representations guided by 3D vision-language aligned representations, and is thus more beneficial to high-level semantic and instance segmentation performances both in terms of data efficiency and open-world recognition capacity. Also, our proposed hierarichical feature alignment also provides more separated feature space, which means that the feature alignment successfully enhances the final instance discrimination capacity.

\noindent
\textbf{Robot Arm Grasping Example.} We have deployed our approach for the task of open-world perception in robot grasping in our extended robotic research work. We use our proposed approach for segmentation and use the ROS 2 Gazebo-based framework to implement the other components of the system, such as kinematics/dynamics modeling, motion planning, low-level control, point cloud-based pose estimation, etc. Our proposed approach has robust performance and decent accuracy in grasping, which demonstrates the potential of our proposed \textit{WS3D++} in industrial manipulation applications in grasping and dropping novel objects beyond the training set.

\vspace{2mm}

\section{Conclusion}
% \vspace{-0.1mm}
% \vspace{-0.56mm} Boundary Prediction Precision grasping a robot 9 6 3 2 8
In this paper, we proposed a general \textit{WS3D++} framework for \textit{open-vocabulary} and \textit{data-efficient} 3D scene parsing. The whole framework involves both the pre-training and the fine-tuning stages. During the pre-training stage, we propose the hierarchical feature alignment strategy to acquire accurate regional 3D-linguistic pairs, thus the performance can be enhanced to a large extent. At the same time, we propose an unsupervised boundary-aware energy-based loss and a novel region-level multi-stage semantic contrastive learning strategy, which are complementary to each other to make the network learn more meaningful and discriminative features from the unlabeled data. The effectiveness of our approach is verified across three diverse large-scale 3D scene understanding benchmarks under various experiment circumstances. Our approach can maximally exploit the unlabeled data to enhance the performance both for 3D point clouds semantic segmentation and instance segmentation, and object detection under various labeling percentages in the limited reconstruction case. Our proposed label-efficient learning framework, termed \textit{WS3D++}, provides conprehensive baselines for future 3D scene parsing methods when the label is inaccessible or limited. The proposed pre-training as well as fine-tuning approach can have a significant boost on the final open-vocabulary and data-efficient semantic scene parsing in term of efficiency, effectiveness, and robustness. 
% \vspace{-1.21mm}
\vspace{-0.11mm}

\vspace{2mm}
\ifCLASSOPTIONcaptionsoff
  \newpage
\fi

% \end{thebibliography}

% or if you just want to reserve a space for a photo:  He received from the Department of Mechanical and Automation Engineering, 
% \vspace{-3mm} 
% \bibliographystyle{Bibliography/IEEEtranTIE}
% \bibliography{Bibliography/IEEEabrv,Bibliography/BIB_xx-TIE-xxxx}\  \small \fontsize{5.0 pt}{ 

% \small
% \bibliographystyle{IEEEtran}
% % \addtolength{\itemsep}{-1.5 em} {10} 10 {1} {16}
% \small
% \bibliography{egbib}
% \small reference  7 \vspace{-1mm} 3 8 
% \vspace{-0.5mm} 10

\vspace{-2.2mm}

\newpage
\bibliographystyle{IEEEtran}

\bibliography{egbib}
\vspace{-1mm}
\vspace{-1.98mm}

\end{document}